\documentclass[10pt,twocolumn,letterpaper]{article}
\usepackage{cvpr}
\usepackage{times}
\usepackage{epsfig}
\usepackage{graphicx}
\usepackage{amsmath}
\usepackage{amssymb}
\usepackage{pifont}

\usepackage{multirow}
\usepackage{xcolor}
\usepackage{enumitem}
\usepackage{colortbl}
\usepackage{booktabs}
\usepackage{mwe}
\usepackage{bm}
\usepackage{dirtytalk}

\newcommand\myparagraph[1]{\textbf{#1} ---}

\graphicspath{ {./images/} }

\def\eg{\emph{e.g}}

\def\ie{\emph{i.e}}
\def\etc{\emph{etc}}

\definecolor{colorh}{rgb}{0.8, 0.8, 0.8}
\definecolor{colorreason}{rgb}{0.30, 0.45, 0.69}
\definecolor{colorbias}{rgb}{0.87, 0.52, 0.32}
\definecolor{colorother}{rgb}{0.34, 0.66, 0.41}

\newcommand{\crossmark}{\ding{55}}

\usepackage[pagebackref=true,breaklinks=true,letterpaper=true,colorlinks,bookmarks=false]{hyperref}

\cvprfinalcopy 

\def\cvprPaperID{987} 

\ifcvprfinal\fi
\begin{document}

\title{Roses are Red, Violets are Blue... But Should VQA expect Them To?}

\author{
	Corentin~Kervadec$^{1,2}$ \quad Grigory~Antipov$^1$ \quad
	Moez~Baccouche$^1$ \quad Christian~Wolf$^2$ \\
	$^1$Orange, Cesson-S\'evign\'e, France. \quad $^2$LIRIS, INSA-Lyon, UMR CNRS 5205, France\\
	{\tt\small corentinkervadec.github.io \{grigory.antipov, moez.baccouche\}@orange.com   christian.wolf@insa-lyon.fr}
}

\maketitle

\begin{abstract}
\noindent
Models for Visual Question Answering (VQA) are notorious for their tendency to rely on dataset biases, as 
the large and unbalanced diversity of questions and concepts involved and tends to prevent models from learning to \say{reason}, leading them to perform \say{educated guesses} instead.
In this paper, we claim that the standard evaluation metric, which consists in measuring the overall in-domain accuracy, is misleading. Since questions and concepts are unbalanced, this tends to favor models which exploit subtle training set statistics.
Alternatively, naively introducing artificial distribution shifts between train and test splits is also not completely satisfying. First, the shifts do not reflect real-world tendencies, resulting in unsuitable models; second, since the shifts are  handcrafted, trained models are  specifically designed for this particular setting, and do not generalize to other configurations.
We propose the GQA-OOD benchmark designed to overcome these concerns: 
we measure and compare accuracy over both rare and frequent question-answer pairs, and argue that the former is better suited to the evaluation of reasoning abilities, which we experimentally validate with models trained to more or less exploit biases. In a large-scale study involving 7 VQA models and 3 bias reduction techniques, we also experimentally demonstrate that these models fail to address questions involving infrequent concepts and provide recommendations for future directions of research.
\end{abstract}

\section{Introduction}
\label{sec:intro}

\begin{figure}[t] \centering
    \includegraphics[width=\linewidth]{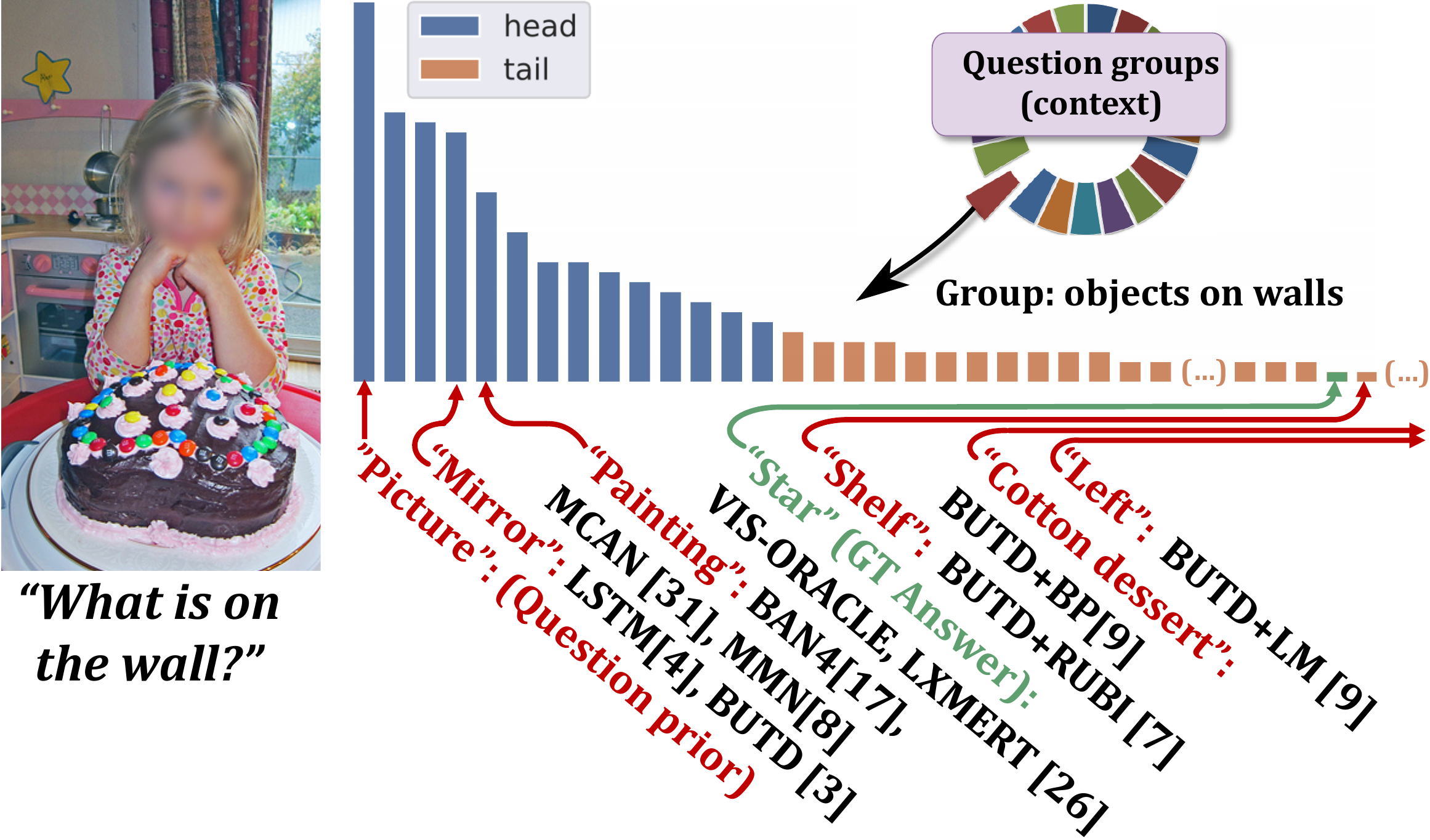}
    \caption{\label{fig:teaser}We address bias exploitation in VQA and propose a new benchmark for Out-Of-Distribution evaluation containing distribution shifts tailored to different question groups with highly imbalanced distributions. A new evaluation metric based on rareness inside each question group, here shown for "objects on walls", is experimentally demonstrated to be less prone to bias exploitation. We show that
    SOTA methods (7 VQA models and 3 bias reduction methods) reproduce biases in training data.}
\end{figure}

\noindent
Visual Question Answering (VQA), i.e. the task of answering a question posed over an image, is often seen as a testbed for the capability of learning-based systems to perform high-level reasoning.
This multimodal 
problem is notorious for its 
diversity, meaning that VQA models are required to learn various high-level general representations of concepts of the physical world as well as their interactions.

Efforts to learn the necessary high-level reasoning from large-scale datasets depend on the absence of harmful biases in the data, which could provide unwanted shortcuts to learning in the form of ``Clever Hans'' effects. Unfortunately, and in spite of recent efforts~\cite{goyal2017making,hudson2019gqa}, most VQA datasets remain very imbalanced. Common concepts are significantly more frequent, e.g the presence of a “\emph{red rose}”, compared to out of context concepts like the presence of a “\emph{zebra in a city}”. This causes 
the tendency of models to overly rely on biases, hindering generalisation~\cite{cadene2019rubi,clark2019don}. 
Despite a general consensus on this diagnostic, systemic evaluations of error distributions are rare. In particular, overall accuracy is still the major, and often unique, metric used to evaluate models and methods,
although it is clearly insufficient.
Several questions remain open.
How is error distributed? Are true positives due to reasoning or to exploitation of bias? What is the prediction accuracy on infrequent vs. frequent concepts?
How can we validate models in Out Of Distribution (OOD)-settings?

In this work we propose a new benchmark and a study of State-Of-The-Art (SOTA) VQA models, 
which allows to precisely answer these questions. 
The proposed new evaluation protocol is complementary to existing ones, but allows a better diagnostic of current VQA performance. In particular, our benchmark can be viewed as an alternative to the VQA-CP~\cite{vqa-cp} dataset, which has lead to mixed results (see \cite{teney2020value} for a recent comprehensive study). Our benchmark comprises (i) a new fine-grained reorganization of the GQA dataset~\cite{hudson2019gqa} introducing distribution shifts in both validation and test sets (see Figure \ref{fig:teaser}-a); (ii) a set of  evaluation metrics; (iii) new evaluation plots illustrating the generalisation behavior of VQA models on different operating points.
The choice of the GQA dataset is motivated by its useful structuring into question groups, which allows to capture biases precisely, to select groups with strong biases and to create distribution shifts tailored to the exact nature of each question (see Figure \ref{fig:teaser}-b). It also makes it possible to analyze how errors are distributed over different associations of concepts according to their frequency in the dataset.

\myparagraph{Our contributions}
(i) We propose and make public\footnote{\url{https://github.com/gqa-ood/GQA-OOD}} a new fine-grained re-organization of the GQA dataset and a set of the respective evaluation metrics allowing to precisely evaluate the reasoning behavior of VQA models and to characterise and visualise their generalisation behavior on different operating points w.r.t distribution shifts.
(ii)    
Compared to competing benchmarks, our dataset features distribution shifts for both, validation and test, allowing to validate models under OOD conditions.
(iii)    
We experimentally evaluate the usefulness of the proposed metric showing its behavior on models trained to, more or less, exploit biases.
(iv) In a large study, we evaluate several recent VQA models and show that they struggle to generalise in OOD conditions; we also test several SOTA bias reduction methods and show that there is still room for improvement in addressing bias in VQA.

\section{Related Work}

\myparagraph{VQA corpuses}
One of the first large-scale datasets was VQA1~\cite{antol2015vqa} with$\sim$76K questions over 25K realistic images. It started a new task, but was soon found to suffer from biases. \cite{goyal2017making} pointed to strong imbalances among the presented answers and proposed the second (improved) version: VQA2.
\cite{johnson2017clevr} introduced the fully synthetic CLEVR dataset, designed to evaluate reasoning capabilities. Its strong point is its detailed and structured annotation. In~\cite{hudson2019gqa}, CLEVR is adapted to real-world images resulting in the automatically created GQA dataset (1.7M questions), offering a better control on dataset statistics.

\myparagraph{Attempts to reduce the bias-dependency}
Despite efforts to design complex architectures, VQA models suffer from significant generalization incapacities~\cite{agrawal2016analyzing}. They tend to answer questions without using the image, and even when they do, they do not always exploit relevant visual regions~\cite{vqahat}.
They tend to overly rely on dataset biases~\cite{hendricks2018women}, and are not able to generalize to unseen distributions~\cite{vqa-cp}. Methods to solve this include \eg ~setting up an adversarial game against a question-only adversary to regularize training~\cite{ramakrishnan2018overcoming}, and a question-only branch in addition to a base model during training to prevent it from learning textual biases ~\cite{cadene2019rubi} (RUBi). At test time, this branch is omitted. \cite{clark2019don} regularize model predictions using question type statistics from the training set. Other attempts force VQA models to attend to the most significant visual regions from humans' perspective \cite{wu2019self, selvaraju2019taking}. While they show promising results when evaluated on unseen distributions~\cite{vqa-cp}, they generally slightly degrade performances on  standard benchmarks, which tend to favor models relying on dataset biases.

\myparagraph{Reinventing VQA evaluation}
Propositions of new evaluation methods for VQA went hand in hand with the design of new models. Early work~\cite{malinowski2014multi} proposed a soft evaluation score based on a lexical database, then replaced by a hard classification score, more prone to favor biased predictions, but easier to use in practice.
The authors of GQA~\cite{hudson2019gqa} proposed several additional metrics, namely: consistency, plausibility, validity, distribution, \etc. They provide insights on  VQA performance, but do not evaluate its ability to predict correct answers in OOD setting, especially when applied to the original balanced GQA dataset. To evaluate the generalization capability of Neural State Machines, \cite{hudson2019learning} proposed interesting splits of GQA, reorganizing the train and validation splits to distinctly evaluate how well the models generalize on visual content and on the linguistic structure of the question.
The authors of~\cite{bahdanauclosure} analyzed generalization to unseen associations of known linguistic structures using their CLOSURE benchmark.
They demonstrated that SOTA models (including those which were explicitly designed to overcome this issue) fail to generalize in these conditions.
This confirms the need of carefully designed benchmarks to evaluate the true capabilities of VQA models. In this context, \cite{vqa-cp} introduced a reorganisation of the VQA2~\cite{goyal2017making} dataset splits, namely VQA-CP2, where the training distribution is made explicitly different from the one in the test. We discuss the differences between VQA-CP2 and our benchmark in detail in Section~\ref{sec:construction}. 

\begin{figure} \centering
    \includegraphics[width=\linewidth]{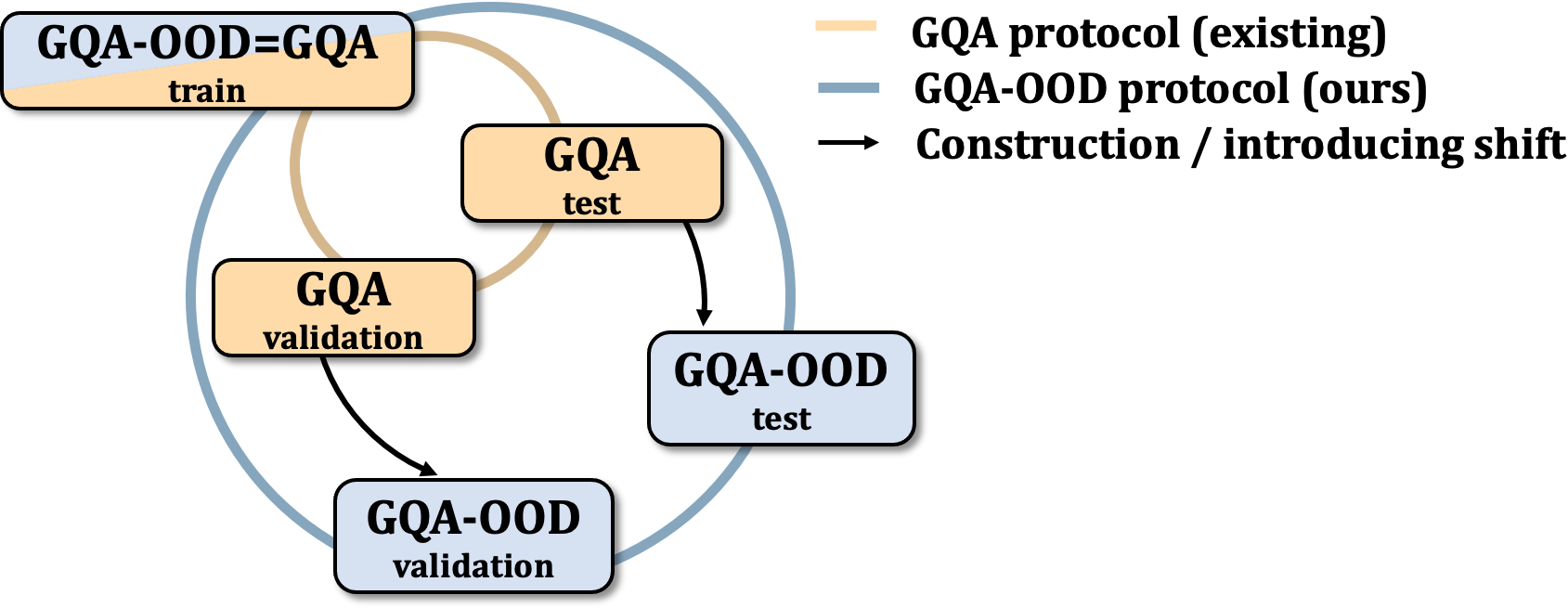}
    \caption{\label{fig:protocol}We re-organize the GQA dataset~\cite{hudson2019gqa} in a fine-grained way: the benchmark contains a distribution shift in validation and test, allowing to validate \emph{and} evaluate in OOD settings.}
\end{figure}

\paragraph{Shortcomings of Out-of-distribution testing}
Recent works have raised some criticizes about OOD evaluation protocols. In particular, \cite{teney2020value} points out several pitfalls observed when evaluating VQA in OOD setting using VQA-CP~\cite{vqa-cp}: 
(1) this protocol does not allow to test a model in both out- and in-distribution settings without having to retrain it on a different set of data;
(2) several works rely of known construction procedures of the OOD test split, \textit{e.g.} the test answers' distribution is the inverse of the train one, coming up with methods specialized to a very particular setup;
and (3) as no validation split is provided, the hyper-parameters are usually selected on the test split.
Although these wrong practices are in part to the responsibility of the model designers, they show us the need to rethink the evaluation protocol.
In parallel, \cite{shrestha2020negative} come up with an interesting negative result while analysing visual grounding bias-reduction methods designed on top of the VQA-CP dataset. These methods~\cite{wu2019self, selvaraju2019taking}, consisting in supervising a VQA model to attend to visual regions which are relevant to a human considering the question, are very efficient on VQA-CP. Surprisingly, \cite{shrestha2020negative} found that simply enforcing the model to attend to random visual regions was at least as much efficient on out- and in-distribution settings. Why was this negative result not observed before? We think that a more profound empirical evaluation of models' behavior would help to better judge and compare the efficiency of new VQA methods. 

These works unveil the need of a better evaluation protocol, allowing to test in both out- and in-distribution settings without falling into ood testing pitfalls. In this paper, we try to provide this evaluation protocol. We complement it with a large experimental study of the generalisation abilities of VQA architectures and de-bias methods in order to better understand theirs pros and cons.


\section{GQA-OOD: a benchmark for OOD settings}
\label{sec:construction}
\noindent
We introduce a new VQA benchmark named \emph{GQA-OOD} designed to evaluate models and algorithms in OOD configurations. 
We here define OOD samples as rare events, in particular measured w.r.t. to a base distribution, \textit{e.g.} a training distribution. 
These rare events might involve concepts which are also present in the training set. Let's for instance consider the question: \textit{`What color is this rose?'}.
If the image represents a rose, then \emph{red} would be a common color, but in an OOD setting, infrequent (correct) test answers would be, for instance, \emph{blue}, requiring models to reason to provide the correct answer. We design a benchmark where this shift is not global but depends on the context. If the context changes, and the flower type is a violet, then a (correct) OOD answer would now be \emph{red} instead of \emph{blue}.

\begin{table}[t] \centering
{ \footnotesize
\setlength{\tabcolsep}{3pt}
\begin{tabular}{llrrl}
\hline\noalign{\smallskip}
Dataset & Split & \#Quest. & \#Groups & \#Imgs \\
\noalign{\smallskip}
\hline
\noalign{\smallskip}
\multirow{2}{*}{GQA-OOD} & val & $51,045$ & $3,849$ & $9,406$\\
 & testdev & $2,796$ & 471 & 388\\
 \hline
\multirow{2}{*}{GQA} & val & $132,062$ & $36,832$ & $10,234$\\
 & testdev & $12,578$ & $7,803$ & 398\\
\hline
\end{tabular}
\\
(a)
\\
\vspace{2mm }
\begin{tabular}{llrrll}
\hline\noalign{\smallskip}
Split & Subset & \#Quest. & \#Groups & \#Imgs \\
\noalign{\smallskip}
\hline
\noalign{\smallskip}
\multirow{2}{*}{GQA-OOD val} & head & $33,882$ & $3,849$ & $8,664$ \\
 & tail & $17,163$ & $3,849$ & $6,632$\\
 \hline
\multirow{2}{*}{GQA-OOD testdev} & head & $1,733$ & 471 & 365\\
 & tail & $1,063$ & 471 & 330 \\
\hline
\end{tabular}
\\
(b)
}
\setlength{\tabcolsep}{1.4pt}
\vspace*{1mm}
\caption{\label{table:stats}Data statistics: (a) GQA-OOD vs. GQA; (b) head vs. tail}
\end{table}

The \emph{GQA-OOD} benchmark consists of a dataset and new evaluation metrics. The dataset itself is based on the existing GQA~\cite{hudson2019gqa} dataset\footnote{We use version 1.2 of the GQA dataset~\cite{hudson2019gqa}.}, which provides  more fine-grained annotations compared to competing VQA2~\cite{goyal2017making} (the questions in GQA have been automatically generated from scene graphs, which allows better control of the context). Figure \ref{fig:protocol} shows how the proposed protocol compares to the existing GQA protocol: the two share the same (existing) training set, but we introduce fine-grained shifts into both the validation and the test sets applying the process further described below.
The shifted subsets have been constructed in $3$ steps: (i) dividing questions into groups according to their contexts; (ii) extracting the most imbalanced question groups considering their answer distributions; (iii) selecting OOD samples among the remaining questions. 

\myparagraph{Question groups}
To structure the process introducing distribution shifts, we use the notion of \textit{local groups} provided in the GQA annotation. They allow to precisely define the type of a question, \eg ~\textit{`What color ...?'}, \textit{`Where is ...?'}, \etc. They also depend on the 
concepts related to the question, \eg ~\textit{`zebra'}, \textit{`violet'}, \etc.
There is a total of $\approx37K$ local groups related to 
$\approx132K$ questions in the GQA validation split.
We use the balanced version of GQA, whose question distribution has been smoothed in order to obtain a more uniform answer distribution. However, this does not impact the imbalanced nature of the dataset, which is often due to real world tendencies, \textit{e.g.} that \textit{`roses are red'}.

\myparagraph{Measuring group imbalance}
We extract a subset of the most imbalanced question groups, as we are interested in evaluating the prediction error specifically in the context, where shifts in distribution are meaningful and strong. We measure balance through Shannon entropy given as 
$
    e(x) = - \sum_{i=0}^d  p(x_i) \log p(x_i),
$
where $p(x_i)$ is the estimated probability of the class $i$.
As entropy depends on the number of answer classes, which is highly variable between different question groups, we normalize entropy w.r.t. the number $d$ of possible answers in the group:
$
    \bar{e}(x) = \frac{e(x)}{\log(d)}, 
$
where $\log(d)$ is equal to the entropy of a uniform distribution of size $d$. Normalized entropy $\bar{e}(x)$  thus measures how close the distribution $p(x)$ is to a  uniform distribution of the same dimension.
Finally, we keep groups with a normalised entropy smaller than a threshold empirically set to $T{=}0.9$. This selects all benchmark's questions, but further work is done in order to select specific answer classes for each group.

\myparagraph{OOD setting and metrics}
We introduce a shift in distribution by selecting a subset of answer classes for each question group according to their frequencies, and introduce three different metrics according to which classes are used for evaluation. All these metrics are defined over the aforementioned imbalanced local groups. Figure~\ref{fig:teaser} illustrates how the subsets are selected using the concrete example answer histogram of question group \textit{objects on walls}.
\begin{itemize}[nosep,itemsep=1mm]
    \item \textbf{Acc-tail}: the accuracy on OOD samples, which are the samples of the tail of the answer class distribution, \textit{i.e.} the rarest answers given the context.
    We define the tail classes as classes $i$ with $|a_i| \le \alpha \mu(a)$, where $|a_i|$ is the number of samples belonging to the class $i$ and $\mu(a)$ is the average sample count for the group.
    We empirically set the parameter $\alpha{=}1.2$, and in Section~\ref{sub:err_dist} we analyze and illustrate the impact of the choice of $\alpha$ on \textit{Acc-tail}.
    Figure~\ref{fig:teaser} provides an example of such a tail question --- we can see that the answer \textit{Star} is rare in this group, therefore it belongs to the tail set like the other answers shown in orange.
    \item \textbf{Acc-head}: the accuracy on the distribution head for each local group, given as difference between the whole group and its tail (blue answers in Figure~\ref{fig:teaser}).
    \item \textbf{Acc-all}: the overall (classical) accuracy over all \emph{GQA-OOD} samples, \textit{i.e.} the in-domain accuracy. In Figure~\ref{fig:teaser}, this corresponds to the blue and orange answers. 
\end{itemize}

Table \ref{table:stats} provides statistics of the proposed benchmark. We also analyzed the nature, distribution and diversity of the questions w.r.t to GQA~\cite{hudson2019gqa}, \textit{c.f.} Appendix.

\begin{figure*}[t] \centering
\begin{tabular}{ccc}
    \begin{minipage}{2cm}
        \includegraphics[width=\linewidth]{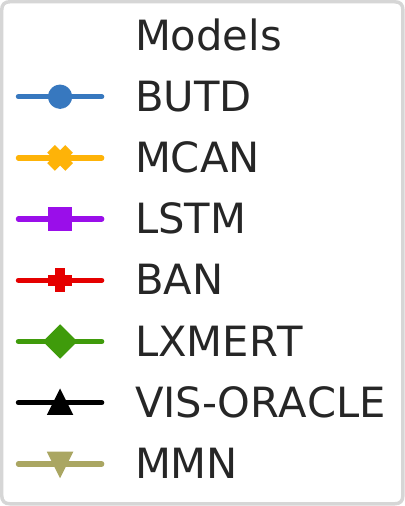}
    \end{minipage}
    &
    \begin{minipage}{7cm}
        \includegraphics[width=\linewidth]{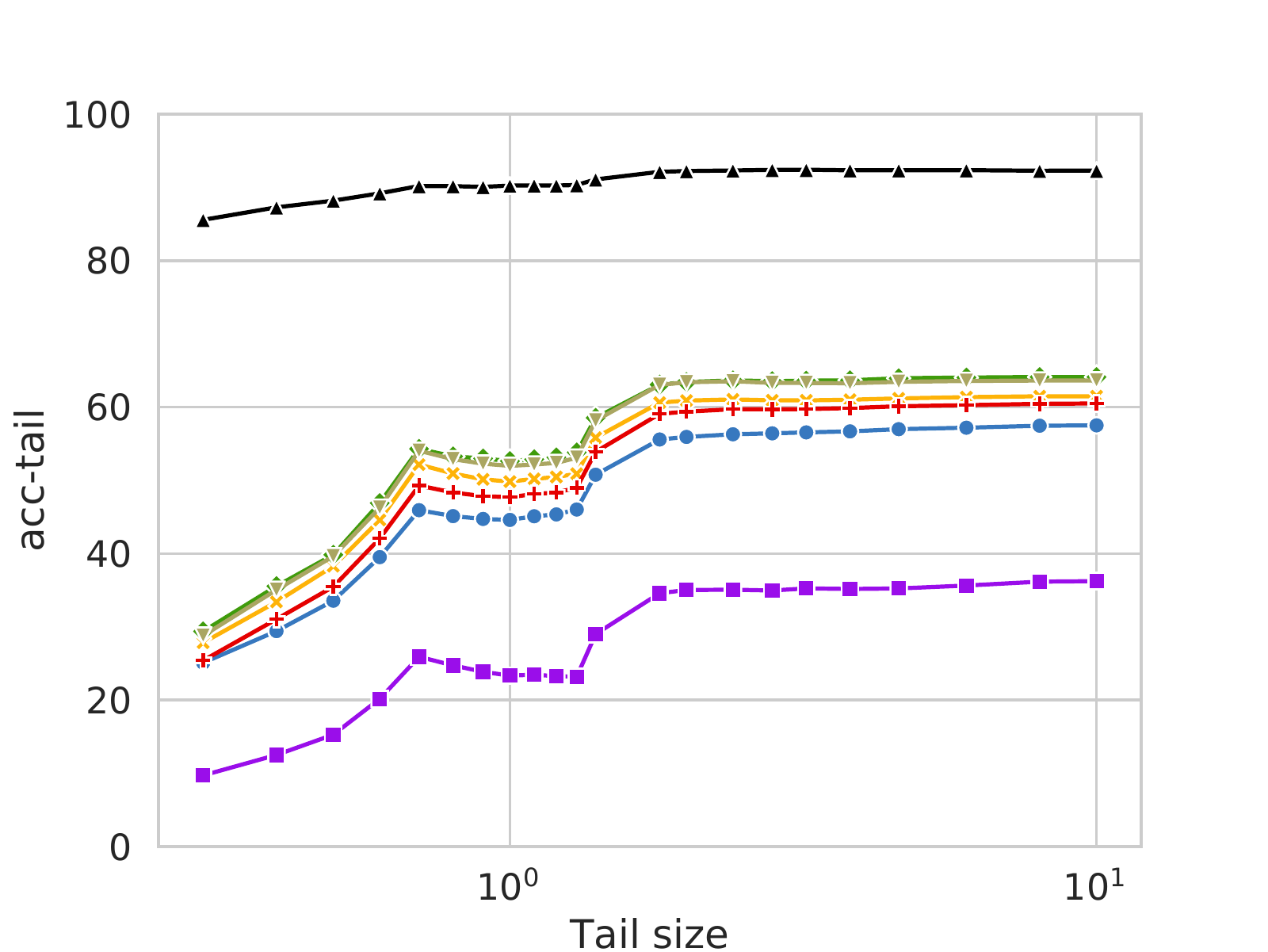}
    \end{minipage}
    &
    \begin{minipage}{7cm}
        \includegraphics[width=\linewidth]{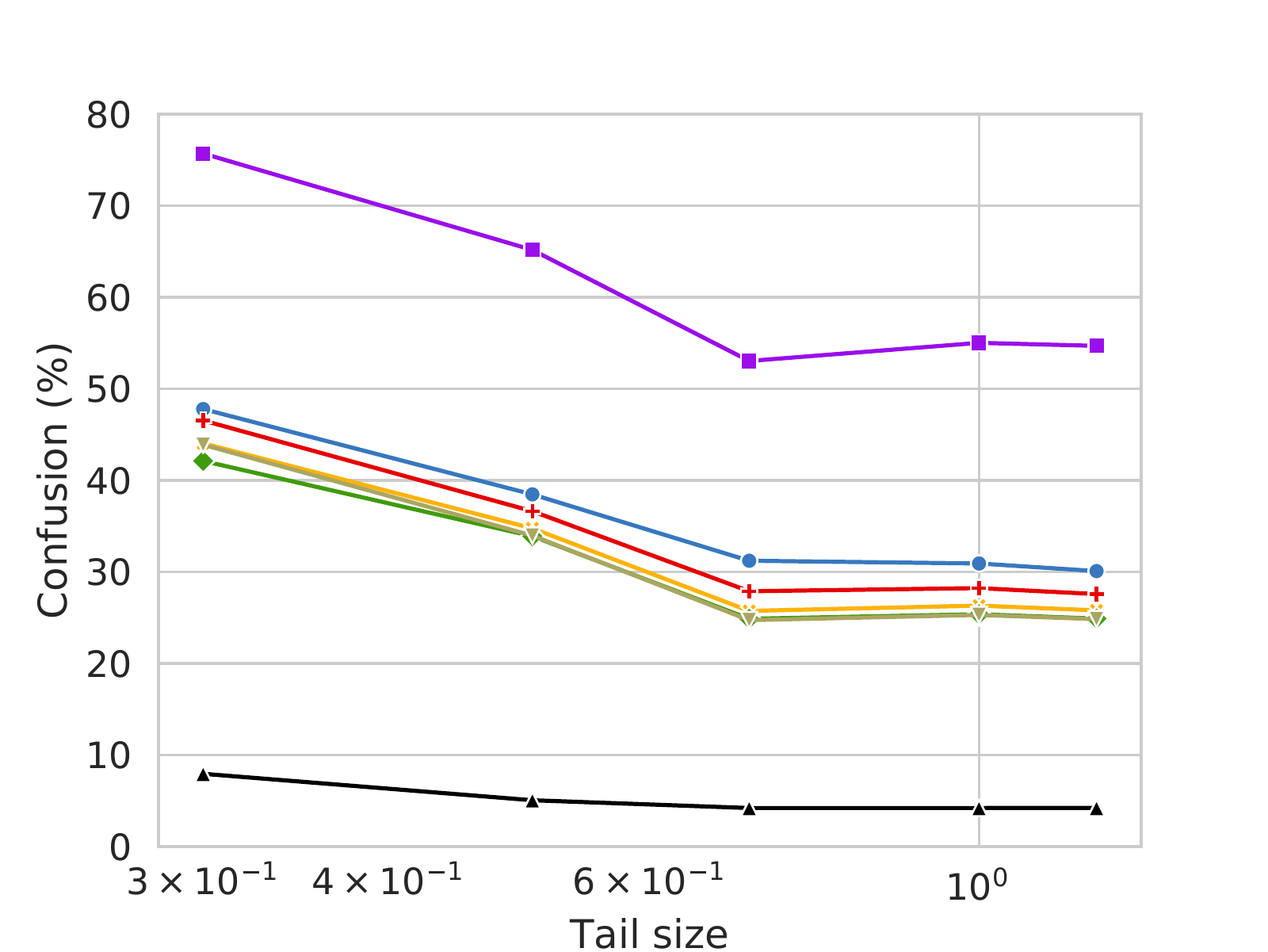}
    \end{minipage}
    \\
    & (a) & (b)
    \\
\end{tabular}
\vspace*{1mm}
\caption{\label{fig:acc_tail_models}Performance (higher is better) (a) and head/tail confusion (lower is better) (b) for different definitions of the tail distribution ($\alpha$ parameter values) on the GQA-OOD benchmark. We compare several VQA \textbf{models}. The x-axis is in log-scale.}
\end{figure*}

\myparagraph{Difference with VQA-CP2}
The VQA-CP2~\cite{vqa-cp} dataset was a first of its kind and paved the way for follow-up work on bias reduction methods in VQA. However, its construction is conceptually different from our work, partially due to the restrictions of the base dataset VQA2 w.r.t. to GQA, but also due to key design choices.
Lacking annotations on group structure in the base dataset, questions are grouped  according to their first words and the ground-truth answer. The shift is created by splitting according to types.
In contrast, our proposed \textit{GQA-OOD} dataset allows fine-grained analysis of the generalization behavior of a VQA model by (i) question group, and via (ii) different metrics corresponding to different amounts of shifts (\textit{acc-tail} vs. \textit{acc-head}) in out- and in-distribution settings, and (iii) even through the possibility of continuous evaluation along different operating points (see Figure \ref{fig:acc_tail_models}).

VQA-CP2 comprises two splits only (train+test), lacking the possibility of validating model hyper-parameters. Most techniques therefore seem to optimize their hyper-parameters on the test split~\cite{cadene2019rubi,clark2019don,ramakrishnan2018overcoming,wu2019self,selvaraju2019taking}, which should be frowned upon,
or, alternatively, validate on a subset of train which does not include a shift~\cite{teney2020Unshuffle}. Our \textit{GQA-OOD} datset contains a validation set with a shift w.r.t. to the train set, which allows to validate hyper-parameters in OOD settings.
Finally, unlike VQA-CP, our proposed dataset requires models to be trained on the existing GQA train split. This forces models to reduce bias in their test results while being exposed to natural tendencies and biases captured in the training corpus, favoring work on bias reduction through methodology instead of through cleaning of training data.

\myparagraph{Limitations}
{
The proposed benchmark is built on the GQA dataset, whose questions have been automatically generated, resulting
in a limited vocabulary (GQA covers $70\%$ of VQA2 answers~\cite{hudson2019gqa}) and a synthetic syntax. While the images are natural and real, one might argue, that the questions are not ``\textit{in the wild}''.
However, the benefits of the synthetic nature of the questions largely out-weight its limitations.
In particular, this offers a better control on the data and excludes unmodelled external knowledge, which leads to a better evaluation of reasoning abilities.
We made the source code publicly available, and we encourage the field to use it to study robustness in OOD settings.}


\section{Experiments}
\label{sec:exp}

\noindent
In our experiments we used several SOTA VQA models and we compared the proposed \emph{GQA-OOD} benchmark to the standard benchmarks VQA2~\cite{goyal2017making}, GQA~\cite{hudson2019gqa} and VQA-CP2~\cite{vqa-cp}. The line-up includes strong models with object-level attention and two Transformer~\cite{vaswani2017attention}-based model, as well as two blind baseline models and a visual oracle.

\myparagraph{Question Prior} this blind baseline returns the most probable answer estimated from training set statistics. Following \cite{vqa-cp,goyal2017making}, 
we use the question types priors when evaluating on VQA-CP and VQA2. For GQA, we use the training global group priors. 
\myparagraph{LSTM~\cite{antol2015vqa}} this blind baseline takes  GloVe embeddings ~\cite{pennington2014glove} and encodes them using an LSTM~\cite{hochreiter1997long} followed by a feed-forward network. Neither of the two blind baselines use the input image.
\myparagraph{BUTD~\cite{anderson2018bottom}} a strong VQA baseline based on object-level attention, in particular, bounding boxes with dense visual feature vectors extracted from the image using an object detector.
\myparagraph{BAN4~\cite{kim2018bilinear}} another object-level attention based model using a bilinear fusion mechanism to models relations between words and objects with multi-hop reasoning. We use the $4$-layers version of BAN.
\myparagraph{MCAN~\cite{yu2019deep}} a Transformer~\cite{vaswani2017attention} architecture designed to model both intra-modality interactions and the inter-modality ones. It allows complex multi-hop reasoning through several stacked self-attention blocks. In our experiments, we use the $6$-layers version of MCAN.
\myparagraph{LXMERT~\cite{tan2019lxmert}} a SOTA approach combining a Transformer architecture (close to the one used in MCAN) with a large scale pre-training procedure. LXMERT is pretrained on rougly 9 millions question-image paires with semi-supervised tasks such as reconstructing masked words and visual objects. It is then finetuned on the GQA dataset.
\myparagraph{MMN~\cite{chen2021meta}} a Meta Module Network for compositional visual reasoning. It is based on a hybrid approach combining neural module networks (NMN) and monolithic architectures (such as MCAN). The former, NMN, is based on hand-crafted neural network program blocs and is supposed to allow a better compositionality and interpretability. The latter, monolithic architecture, performs its operations in a latent space and has been shown to be experimentally more efficient. NMN tries to combine the best of both worlds. 
\myparagraph{VIS-ORACLE} a model with a perfect sight, \textit{i.e.} taking as input the question and a set of ground truth objects directly taken from the annotation of GQA\footnote{As GT annotations (scene-graphs) are only available for the train and validation split, we do not evaluate VIS-ORACLE on the testdev split.}. It allows to evaluate the performance of a model without the imperfection of the visual extractor. Is is based on a Transformer architecture similar to the one used for LXMERT with smaller capacity.

For training details of all models we refer to the Appendix.
Models evaluated on \emph{GQA-OOD} were trained on the training set of GQA (balanced) and validated on the validation split of \emph{GQA-OOD}, unless otherwise stated. When available, we provide the standard deviation computed over at least four different seeds.

\setlength{\tabcolsep}{4pt}
\begin{table}[t] \centering
{\footnotesize
\begin{tabular}{lllll}
\hline\noalign{\smallskip}
Model& 
\multicolumn{1}{c}{Baseline benchm.}  & 
\multicolumn{3}{c}{Proposed benchmark (Acc-tail)} 
\\ 
& 
\multicolumn{1}{c}{Tot. Acc.} & 
$\alpha{=}1.2$ & $\alpha{=}0.5$ & $\alpha{=}0.3$ 
\\
\noalign{\smallskip}
\hline
\noalign{\smallskip}
BUTD~\cite{anderson2018bottom} {+} \textit{bal} & 60.7\tiny{$\pm0.4$} & 45.4\tiny{$\pm0.3$}& 33.8\tiny{$\pm0.5$} & 24.6\tiny{$\pm0.5$}\\
BUTD~\cite{anderson2018bottom} {+} \textit{all} & 59.8\tiny{$\pm0.1$} & 41.9\tiny{$\pm0.1$}& 29.5\tiny{$\pm0.3$} & 18.3\tiny{$\pm0.6$}\\
\hline
$\Delta$ (relative): & -1.4\% & -7.7\% & -12.9\% & -25.7\%\\
\hline
\end{tabular}
}
\vspace*{1mm}
\caption{\label{table:train_bias}We compare two different VQA models based on BUTD~\cite{anderson2018bottom}, one of which has been trained on a split known to be biased (BUTD~\cite{anderson2018bottom}{+}\textit{all}), and evaluate the proposed metric's capacity to detect this bias. All scores in \% on the GQA-OOD val split. }
\end{table}
\setlength{\tabcolsep}{1.4pt}

\subsection{Evaluation of the proposed metric}

\noindent
We believe that a good evaluation metric satisfies at least two properties: it is easy to interpret, and it provides an estimate for the quality targeted by the evaluation. We argued above on the merits of our proposed tail accuracy (Acc-tail) as a way of estimating VQA performance less influenced by bias. In what follows, we complete this by an experimental validation of the metric. To this end, we compared two different VQA models, one of which has been trained in a way known to be biased.
In particular, we trained BUTD~\cite{anderson2018bottom}, known to capture training set biases~\cite{vqa-cp}, on the GQA and GQA-OOD validation splits. The first version, BUTD{+}\textit{bal}, is trained on the widely used balanced training set of GQA, which we had also used for all other experiments in this paper. This training set had been created by smoothing the question distribution in order to mitigate dataset biases~\cite{hudson2019gqa}. The second one variant, BUTD{+}\textit{all}, is trained on the raw and unbalanced GQA training set, which leads to  more spurious biases than the balanced version. As the unbalanced set is ten times bigger than the balanced one, we split it in ten subsets and provide the average score.

Results are given in Table~\ref{table:train_bias}, comparing two different metrics, namely the classical total accuracy and our GQA-OOD \textit{acc-tail} metric, with three different values for the $\alpha$ hyper-parameter. First, we observe that the two versions of BUTD obtain similar scores on GQA overall --- the relative difference is only $1.4\%$. This is not a surprise, the classical metric is influenced by biases. As expected, the two VQA models behave differently on our proposed  \textit{acc-tail} metric: the model trained on the unbalanced training set is outperformed by the balanced one by a large margin. Moreover, the score difference increases with decreasing $\alpha$, (\textit{i.e.} when the metric focuses on the rarer and rarer question-answer pairs, providing valuable evidence that \textit{acc-tail} is indeed well suited for measuring VQA performance undisturbed by bias dependencies.

\subsection{Analysis of VQA model error distributions}
\label{sub:err_dist}

\setlength{\tabcolsep}{4pt}
\begin{table*}[t] \centering
\vspace*{1mm}
\begin{tabular}{cp{5mm}c}
\begin{minipage}{8cm}
{\footnotesize
\centering
\begin{tabular}{lclllr}
\hline\noalign{\smallskip}
Model & Uses image & acc-all & acc-tail & acc-head & $\Delta$\\
\noalign{\smallskip}
\hline
\noalign{\smallskip}
Quest. Prior &\crossmark& 21.6 & 17.8 & 24.1 & 35.4 \\
LSTM~\cite{antol2015vqa} & \crossmark& 30.7 & 24.0 & 34.8 & 45.0 \\
BUTD~\cite{anderson2018bottom} &\checkmark& 46.4\tiny{$\pm1.1$} & 42.1\tiny{$\pm0.9$} & 49.1\tiny{$\pm1.1$} & 16.6 \\
MCAN~\cite{yu2019deep} &\checkmark& 50.8\tiny{$\pm0.4$} & 46.5\tiny{$\pm0.5$} & 53.4\tiny{$\pm0.6$} & 14.8 \\
BAN4~\cite{kim2018bilinear} &\checkmark& 50.2\tiny{$\pm0.7$} & 47.2\tiny{$\pm0.5$} & 51.9\tiny{$\pm1.0$} & \textbf{9.9} \\
MMN~\cite{chen2021meta} &\checkmark& 52.7 & 48.0 & 55.5 & 15.6 \\
LXMERT~\cite{tan2019lxmert} &\checkmark& \textbf{54.6} & \textbf{49.8} & \textbf{57.7} & 15.9 \\
\hline
\\
\end{tabular}
}
\end{minipage}
&
&
\begin{minipage}{7cm}
{\footnotesize
\begin{tabular}{llllr}
\hline\noalign{\smallskip}
Technique & acc-all & acc-tail & acc-head & $\Delta$\\
\noalign{\smallskip}
\hline
\noalign{\smallskip}
BUTD~\cite{anderson2018bottom} & 46.4\tiny{$\pm1.1$} & \textbf{42.1}\tiny{$\pm0.9$}& 49.1\tiny{$\pm1.1$} & 16.6 \\
+RUBi+QB & \textbf{46.7}\tiny{$\pm1.3$} & \textbf{42.1}\tiny{$\pm1.0$} & \textbf{49.4}\tiny{$\pm1.5$} & 17.3 \\
+RUBi~\cite{cadene2019rubi} & 38.8\tiny{$\pm2.4$} & 35.7\tiny{$\pm2.3$} & 40.8\tiny{$\pm2.7$} & 14.3\\
+LM~\cite{clark2019don} & 34.5\tiny{$\pm0.7$} & 32.2\tiny{$\pm1.2$} & 35.9\tiny{$\pm1.2$} & \textbf{11.5} \\
+BP~\cite{clark2019don} & 33.1\tiny{$\pm0.4$} & 30.8\tiny{$\pm1.0$} & 34.5\tiny{$\pm0.5$} & 12.0 \\
\hline
\end{tabular}
}
\end{minipage}
\\
(a) && (b)
\\
\end{tabular}
\vspace*{1mm}
\caption{\label{table:distribution_models}Comparison of several VQA methods on the GQA-OOD testdev split. \textit{Acc-tail}: OOD settings, \textit{Acc-head}: accuracy on most probable answers (given context), scores in \%. (a) different models; (b) BUTD~\cite{anderson2018bottom} model combined with different bias reduction techniques.}
\end{table*}
\setlength{\tabcolsep}{1.4pt}

\noindent
The \emph{GQA-OOD} benchmark allows us to perform an analysis of the error prediction distributions for various VQA models as shown in Table~\ref{table:distribution_models}.
We provide the three metrics introduced in Section~\ref{sec:construction}: \textit{acc-tail}, \textit{acc-head} and \textit{acc-all}.
We also measure the difference $\Delta\textit{(tail,head)}=\frac{\textit{acc-head} - \textit{acc-tail}}{\textit{acc-tail}}$ to illustrate how much is the error prediction imbalanced between frequent and rare answers.

\myparagraph{Models fail on rare question-answer pairs}
We can see that that VQA models (dramatically) fail to generalize on infrequent association of concepts.
The two blind models (Question Prior and LSTM in Table~\ref{table:distribution_models}-a) obtain the highest gap between \textit{acc-tail} and \textit{acc-head}, explained by the fact that they uniquely rely on question biases. The $\Delta$ score indicates that BUTD, MMN, MCAN, BAN4 and LXMERT also struggle (in a lesser extent) to generalize on the less frequent question-answer pairs.
Nevertheless, we observe that the Transformer-based architecture combined with large-scale BERT training, LXMERT, outperforms all models on the \textit{acc-tail} metric, confirming its superiority.
{This is corroborated by \cite{hendricks2018women}, who show that pretrained Transformers improve OOD robustness in NLP.}

In contrast to our proposed  \textit{acc-tail} metric, the metric \textit{acc-all}, close to the standard VQA metric, does not reflect the true model’s performances, since it is mechanically increased by the high scores obtained on the most frequent question-answers.
This confirms the need of a two-in-one evaluation: measuring the out- and in-distribution performance scores, as we propose.

\begin{figure}[t] \centering
\begin{tabular}{cc}
    \begin{minipage}{4.1cm}
        \includegraphics[width=\linewidth]{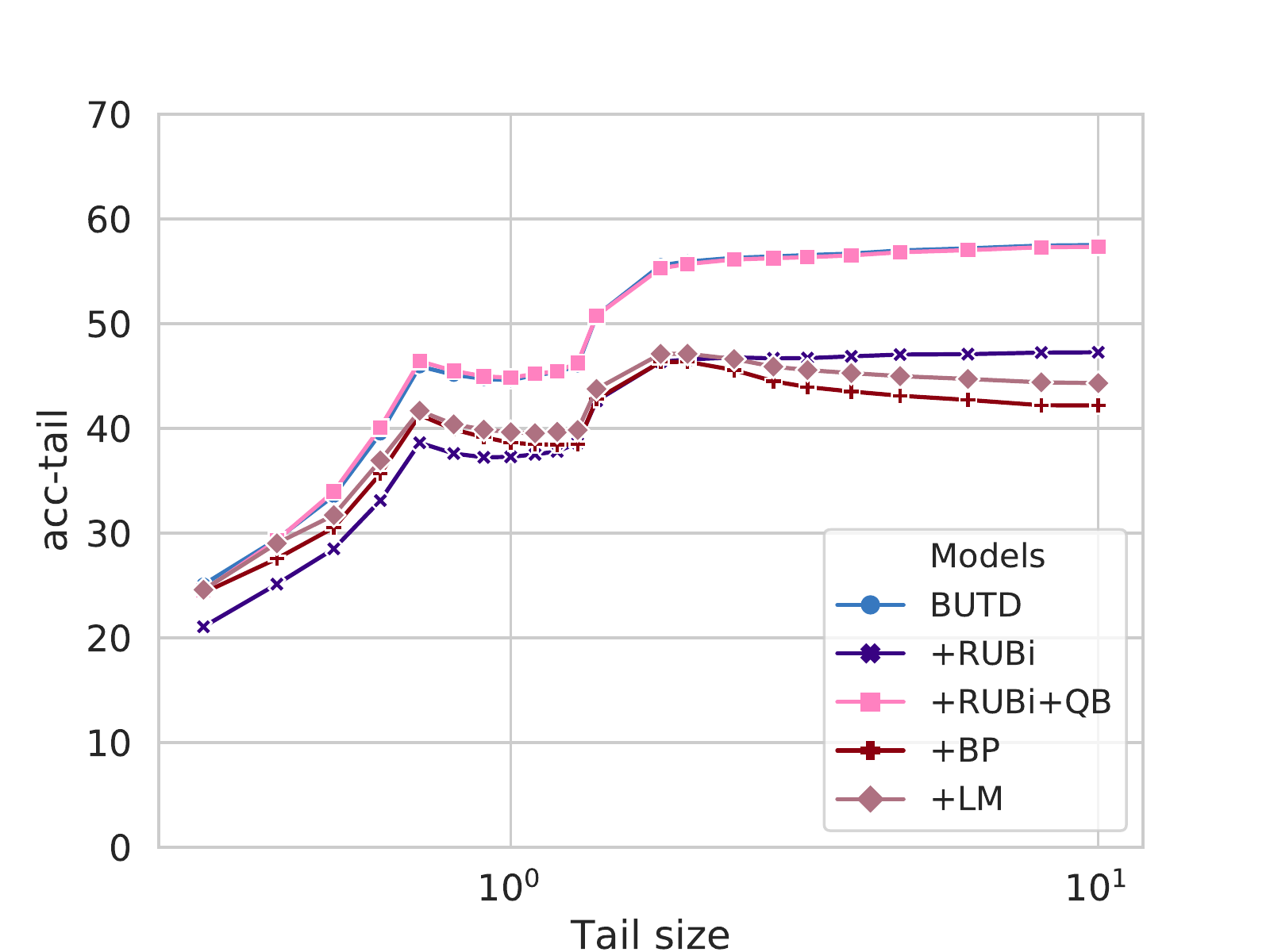}
    \end{minipage}
    &
    \begin{minipage}{4.1cm}
        \includegraphics[width=\linewidth]{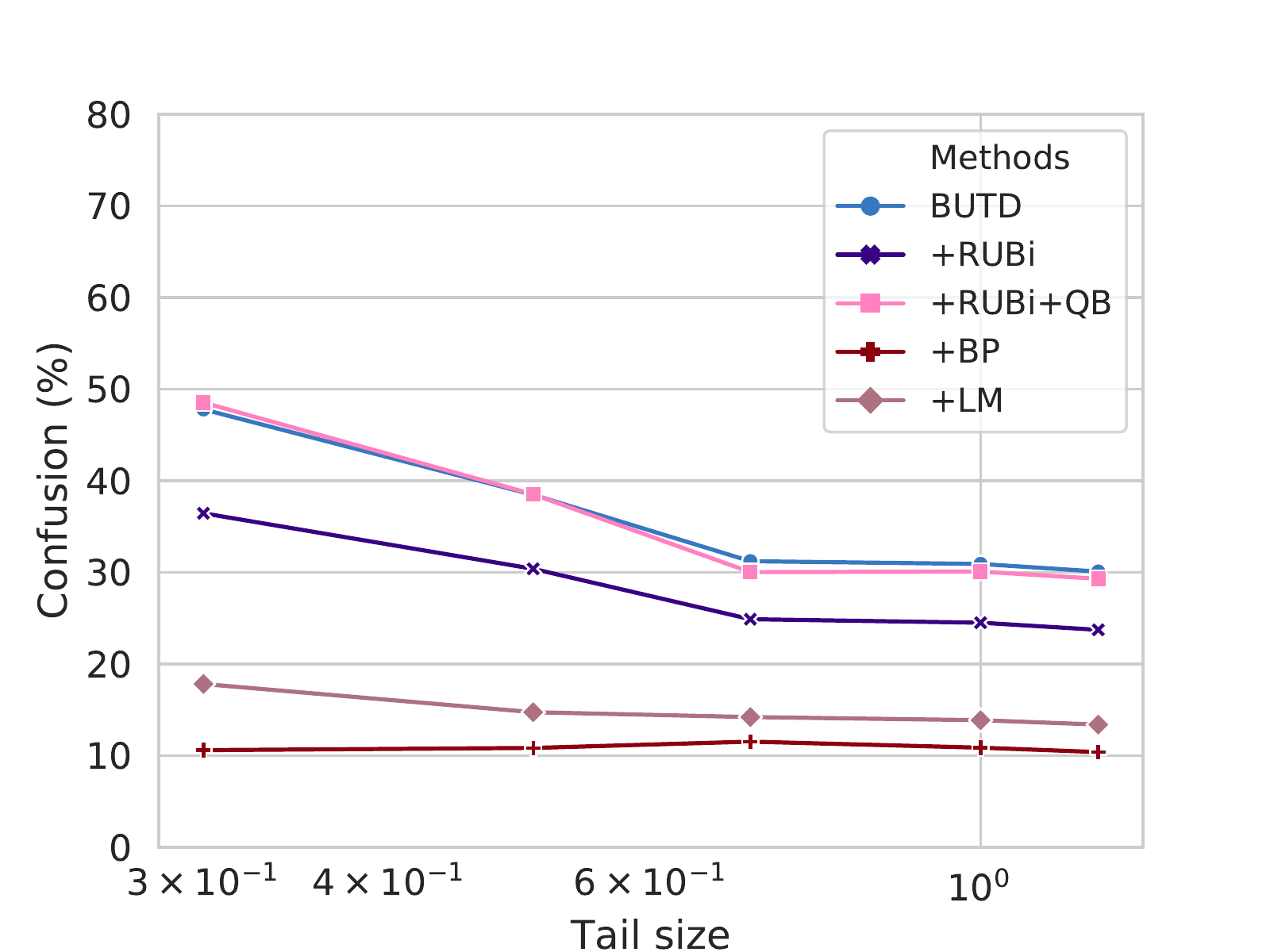}
    \end{minipage}
    \\
    (a) & (b)
    \\
\end{tabular}
\caption{\label{fig:acc_tail_methods}Acc-tail performance and head/tail confusion, as in Fig.  \ref{fig:acc_tail_models}), but for different bias-reduction methods on top of BUTD~\cite{anderson2018bottom}.}
\end{figure}

\myparagraph{Visualising the generalisation behavior}
The definition of what constitutes a ``rare'' answer, i.e. the size of the tail, depends on the parameter $\alpha$. In Figure~\ref{fig:acc_tail_models}-a, we analyze how VQA model prediction errors (\emph{acc-tail}) depend on this definition, i.e. how models behave w.r.t. to questions whose answers are more and more rare. Increasing $\alpha$ increases the tail --- in the extreme case it is equal to the whole distribution (right side of the plot). With small $\alpha$, only the most infrequent question-answer pairs are evaluated (left side of the plot).
All models follow the same dynamic: starting from a tail size which represents roughly half of the question-answer pairs, tail accuracy starts to linearly decrease until reaching a dramatically low score (about $30$ pts lower than the overall accuracy). 
An exception is VIS-ORACLE: its dynamics is nearly flat, prediction error is almost decorrelated from answer rareness. This provides evidence that a model using perfect visual input is able to learn reasoning with significantly decreased dependency on dataset biases.

We complement this analysis by measuring the confusion between \textit{head} and \textit{tail} as a function of $\alpha$, shown in Figure~\ref{fig:acc_tail_models}-b, which provides insights on the causes of the generalisation failure observed in Figure~\ref{fig:acc_tail_models}-a. 
This confusion corresponds to the proportion of questions where the model predicts a \textit{head} answer with a  \textit{tail} GT answer\footnote{More information is available in the appendix.}. For $\alpha{=}1.2$, LXMERT confuses answers for $25\%$ of questions, which increases up to $42\%$ for $\alpha{=}0.3$. Similar behavior is observed for the other models, but interestingly \emph{not} for VIS-ORACLE, where curve is nearly flat, again providing evidence for a low dependency on statistical biases in the training set. 
{As a side note, we found in \cite{kervadec2021how} that initialising LXMERT weights with VIS-ORACLE allows to boost the accuracy on \emph{acc-tail}.}

\begin{figure*}[t] \centering
\setlength{\tabcolsep}{1pt}
\begin{tabular}{cp{2mm}c}
    \begin{minipage}{7cm}
    {\small
    \begin{tabular}{|c|p{0.9cm}|p{0.9cm}|p{0.9cm}|p{0.9cm}|p{0.9cm}|p{0.9cm}|}
	    \hline  	
	    \cellcolor{colorh}
	    & \multicolumn{2}{|c|}{\cellcolor{colorh}\textbf{Head}} & 
	      \multicolumn{2}{|c|}{\cellcolor{colorh}\textbf{Borderline}} & \multicolumn{2}{|c|}{\cellcolor{colorh}\textbf{Tail}} \\
	      \hline
	      \cellcolor{colorh}
	      & \multicolumn{1}{|c|}{\cellcolor{colorh}\textbf{C}}  
	      & \multicolumn{1}{|c|}{\cellcolor{colorh}\textbf{W}}
	      & \multicolumn{1}{|c|}{\cellcolor{colorh}\textbf{C}}  
	      & \multicolumn{1}{|c|}{\cellcolor{colorh}\textbf{W}}
	      & \multicolumn{1}{|c|}{\cellcolor{colorh}\textbf{C}}  
	      & \multicolumn{1}{|c|}{\cellcolor{colorh}\textbf{W}}\\
	    \hline
	    \cellcolor{colorh}\textbf{Head} &
	    \multicolumn{1}{|r|}{30.0\%} &
	    \multicolumn{1}{|r|}{9.6\%}&
	    \multicolumn{1}{|r|}{0.0\%}&
	    \multicolumn{1}{|r|}{3.1\%}&
	    \multicolumn{1}{|r|}{0.0\%}&
	    \multicolumn{1}{|r|}{\cellcolor{colorbias}5.3\%}\\
	    \hline
	    \cellcolor{colorh}\textbf{Borderline} &
	    \multicolumn{1}{|r|}{0.0\%} &
	    \multicolumn{1}{|r|}{5.5\%}&
	    \multicolumn{1}{|r|}{6.3\%}&
	    \multicolumn{1}{|r|}{2.1\%}&
	    \multicolumn{1}{|r|}{0.0\%}&
	    \multicolumn{1}{|r|}{\cellcolor{colorother}3.2\%}\\
	    \hline
	    \cellcolor{colorh}\textbf{Tail} & 
	    \multicolumn{1}{|r|}{0.0\%} &
	    \multicolumn{1}{|r|}{5.0\%}&
	    \multicolumn{1}{|r|}{0.0\%}&
	    \multicolumn{1}{|r|}{0.8\%}&
	    \multicolumn{1}{|r|}{\cellcolor{colorreason}11.6\%}&
	    \multicolumn{1}{|r|}{\cellcolor{colorother}1.3\%}\\
	    \hline
	    \end{tabular}
	    }
	    \\
	    \\
	   {\footnotesize
	   \emph{
	    C=Correct, W=Wrong \\
	    Rows=predicted labels, columns=GT labels \\
	    {\textcolor{colorreason}{Blue=Model is estimated to reason}} \\
	    {\textcolor{colorbias}{Orange=Model is estimated to exploit bias}} \\
	    {\textcolor{colorother}{Green=Unknown label}}
	   } 
	    }
    \end{minipage}
    &
    &
    \begin{minipage}{6cm}
    \includegraphics[width=\linewidth]{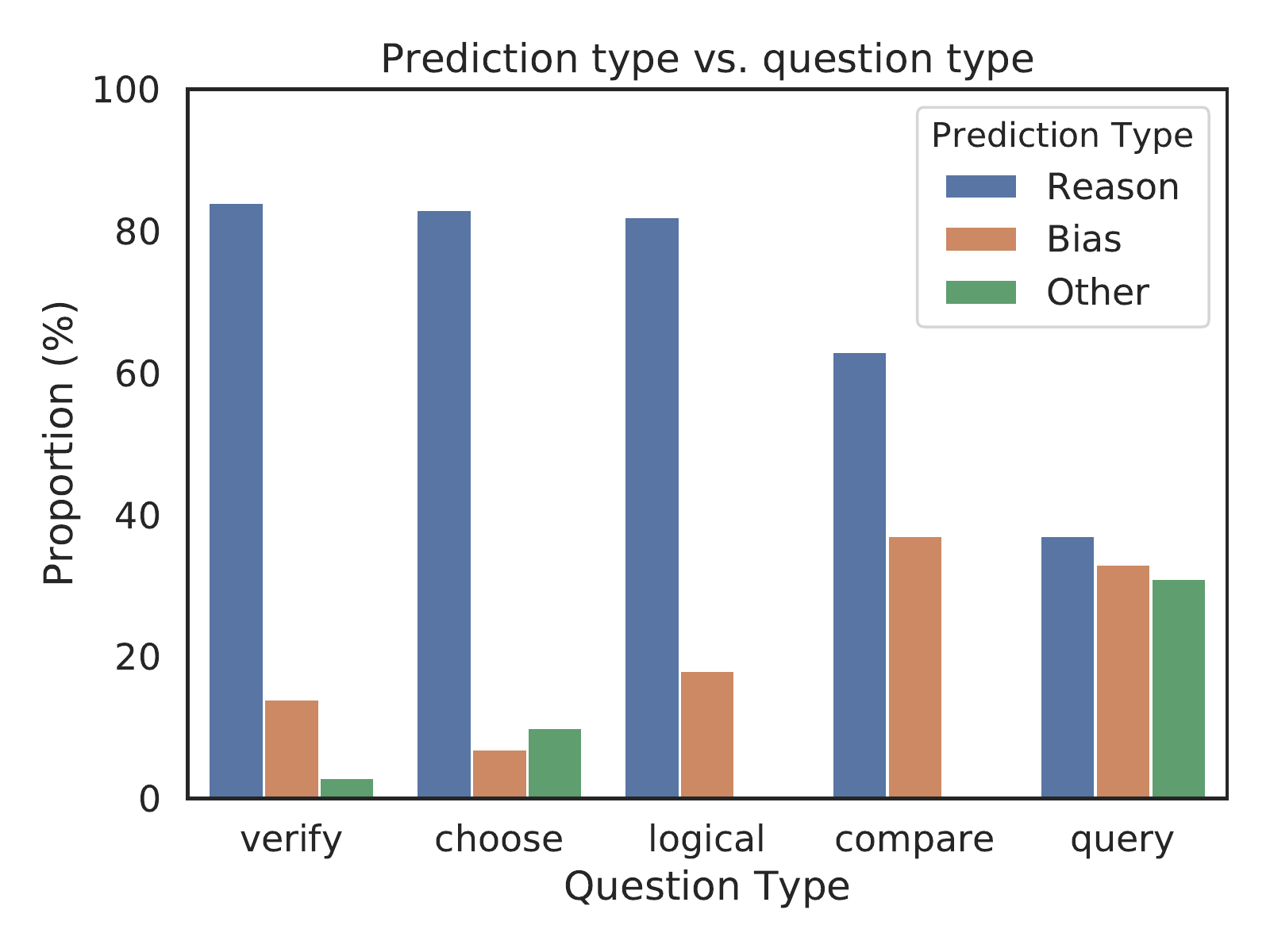}
    \end{minipage}
    \\
    (a) && (b) 
    \\
\end{tabular}
\caption{\label{fig:reasoning_label}We estimate ``\emph{reasoning labels}'': the model is estimated to \textit{reason}, when it correctly predicts an answer rare in GT and rare in prediction; it is considered \textit{biased}, when it wrongly predicts an answer, which is \emph{frequent} in its prediction and \emph{rare} in GT. All values are computed over the GQA-OOD validation split. The matrix (a) shows the joint distribution of predicted and GT classes.
(b): Distribution the estimated reasoning labels over the GQA~\cite{hudson2019gqa} question types for the LXMERT~\cite{tan2019lxmert} model. The model often predicts a biased answer on the \textit{query} and \textit{compare} questions while there is evidence that it may reason on \textit{verify}, \textit{choose} and \textit{logical} questions.}
\setlength{\tabcolsep}{1.4pt}
\end{figure*}

\myparagraph{Exploiting biases vs. ``reasoning''}
It is difficult to assess, whether a model reasons or not, in particular since the term ``reasoning'' has various different definitions. Referring to \cite{bottou2014machine}, \say{\emph{a plausible definition of ‘reasoning’ could be ‘algebraically manipulating previously acquired knowledge in order to answer a new question’}}. In the context of VQA, we could interpret this as \say{\emph{algebraically manipulating words and visual objects to answer a new question}}. With this definition, using statistics biases cannot be considered reasoning, but should rather be denoted as \say{educated guesses}~\cite{hudson2019gqa} or \textit{biased} answers. Using the proposed GQA-OOD benchmark, we explore the estimation of three reasoning labels qualifying the mode of operation a model uses for a given input: \textit{bias}, \textit{reason} and \textit{other/unknown}. In absence of GT information, we propose to estimate these labels from proxy rules:
a VQA model is estimated to \textit{reason}, when it correctly predicts an answer, which is rare in GT and rare in prediction; it is considered \textit{biased}, when it wrongly predicts an answer, which is \emph{frequent} in its prediction and \emph{rare} in GT. 

Figure~\ref{fig:reasoning_label}-a shows the calculation of these labels based on the distribution of the \emph{head} and \emph{tail} labels of each answer in the predictions (rows) and GT (columns) for LXMERT~\cite{tan2019lxmert} on the validation split of GQA-OOD. We add a \textit{borderline} label representing the fuzzy frontier between reasoning and bias exploitation\footnote{\textit{head}: $\alpha{>}1.2$, \textit{borderline}: $0.7{<}\alpha{<}1.2$, \textit{tail}: $\alpha{<}0.7$.}. 
In Figure~\ref{fig:reasoning_label}-b, we show the distribution of these reasoning labels over the different GQA structural question types~\cite{hudson2019gqa}: \textit{verify}, \textit{choose}, \textit{compare} and \textit{query}. We observe that LXMERT seems to \say{reason} on the \textit{verify}, \textit{choose} and \textit{logical} questions, which are binary questions, while \textit{compare}\footnote{only $1\%$ of the tail questions are typed as \textit{compare}.} and \textit{query} questions are the most prone to bias exploitation. From this, we conclude that future efforts on improvements of model capacities to answer open questions (\textit{e.g} typed as \textit{query}) should be particular fruitful.

\subsection{Re-evaluating bias-reduction methods}
\noindent
We use the proposed benchmark to re-evaluate several bias-reduction methods, which have been initially designed on the VQA-CP~\cite{vqa-cp} dataset. As these methods were designed to be model-agnostic~\cite{cadene2019rubi,clark2019don}, we use them together with the BUTD~\cite{anderson2018bottom} architecture:

\myparagraph{RUBi~\cite{cadene2019rubi}} adds  a question-only branch to the base model during  training to prevent it from learning question biases. This branch is omitted during evaluation. To better analyze bias dependencies, we also study a modified version of RUBi, which we refer to as \textbf{RUBi+QB} below. In this variant, the question-only branch is kept during evaluation.
\myparagraph{BP~\cite{clark2019don}} is similar to RUBi but differs by directly taking training set statistics to infer question type biases during  training\footnote{VQA2: biases are over question types; GQA: local groups.}. The question type biases are fused with the base model predictions using a product of experts~\cite{clark2019don}, and removed during testing.
\myparagraph{LM~\cite{clark2019don}} is an improved version of BP~\cite{clark2019don}. In this version, the question bias is dynamically weighted by the base model in order to control its influence. In the original setup, an entropy penalty is added to the loss to prevent the model to ignore the bias. Nevertheless, when training on GQA, we obtain better results without this penalty (see appendix for details).

Surprisingly, none of the three bias-reduction methods succeed to improve \textit{acc-tail} (\textit{cf.} Table~\ref{table:distribution_models}-b), they even deteriorate \textit{acc-head}. This is unexpected as they have been designed to overcome the dependency on question type biases.  For further analysis, we evaluate RUBi while keeping the question-only branch during testing (RUBi+QB). As expected, it outperforms RUBi on \textit{acc-head}, indicating it has better captured frequent patterns. However, it also outperforms RUBi on the OOD settings, demonstrating that preventing from learning frequent patterns does not necessarily increase performances on rare samples.

We provide a visualization of the generalisation behavior on bias-reduction methods in Figure~\ref{fig:acc_tail_methods}-a. For BP~\cite{clark2019don}, LM~\cite{clark2019don} and, to a lesser extent, RUBi~\cite{cadene2019rubi}, we observe that the right side of the curve has flattened, indicating that overall accuracy, dominated by frequent question-answer pairs, has been reduced by  bias-reduction.
The left side of the curve, however, corresponding to rare samples, remains almost unchanged, revealing that these methods have somewhat succeeded in preventing the base model from learning dataset biases. As a comparison, the LSTM model in Figure~\ref{fig:acc_tail_models}-a performs worse than BUTD but conserves the same frequent/rare imbalance.
We observe that RUBi+QB responds the same way as BUTD, confirming the effect of bias-reduction; looking at \textit{head/tail} confusion in Figure~\ref{fig:acc_tail_methods}-b, the result is even more pronounced. 
In short, we demonstrate the effectiveness of bias reduction methods in preventing the base model from learning salient properties of the training set, and occasionally reducing the dependency toward dataset biases. However, this does not necessarily help the model to learn the subtle distributions, required for generalization and for learning to reason.

\setlength{\tabcolsep}{4pt}
\begin{table}[t] \centering
{\footnotesize
\label{table:bench}
\centering
\begin{tabular}{llllll}
\hline\noalign{\smallskip}
\multirow{2}{*}{Model} & VQA2 & \multicolumn{2}{c}{GQA} & VQA-CP2 & GQA-OOD \\
& overall & overall & dist. & overall & acc-tail \\
\noalign{\smallskip}
\hline
\noalign{\smallskip}
Q. Prior                  & 32.1 & \textit{27.0} & \textit{55.6} & 8.8 & \textit{17.8}\\
LSTM~\cite{antol2015vqa}        & 43.0 & \textit{39.1} & \textit{3.6} & \textit{22.1} & \textit{24.0}\\
BUTD~\cite{anderson2018bottom}  & 63.5 & \textit{51.6}\tiny{$\pm0.3$} & \textit{1.8} & 40.1 & \textit{42.1}\tiny{$\pm0.9$}\\
MCAN~\cite{yu2019deep}          & \textbf{\textit{66.1}} & \textit{56.3}\tiny{$\pm0.2$} & \textit{1.6} & \textbf{\textit{42.5}} & \textit{46.5}\tiny{$\pm0.5$}\\
BAN4~\cite{kim2018bilinear}  & 65.9 & \textit{54.7}\tiny{$\pm0.4$} & \textit{1.6} & 40.7 & \textit{47.2}\tiny{$\pm0.5$}\\
MMN~\cite{chen2021meta}         & - & \textbf{59.6} & 1.8 & - & \textit{48.0}\\
LXMERT~\cite{tan2019lxmert}     & \textbf{69.9} & \textbf{\textit{59.6}} & \textbf{\textit{1.5}} & - & \textbf{\textit{49.8}}\\
\hline
BUTD~\cite{anderson2018bottom}  & \textbf{63.5} & \textit{51.6}\tiny{$\pm0.3$} & \textit{1.8} & 40.1 & \textbf{\textit{42.1}}\tiny{$\pm0.9$}\\
+RUBi+QB                       & - & \textbf{\textit{51.9}}\tiny{$\pm1.1$} & \textbf{\textit{1.7}} & \textbf{\textit{47.6}}\tiny{$\pm3.7$} & \textbf{\textit{42.1}}\tiny{$\pm1.0$} \\
+RUBi~\cite{cadene2019rubi}     & 61.2 & \textit{43.6}\tiny{$\pm2.0$} & \textit{1.9} & 44.2 & \textit{35.7}\tiny{$\pm2.3$}\\
+LM~\cite{clark2019don}        & 56.4  & \textit{39.7}\tiny{$\pm0.7$} & \textit{2.1} & 52.0 & \textit{32.2}\tiny{$\pm1.2$}\\
+BP~\cite{clark2019don}         & 63.2  & \textit{39.6}\tiny{$\pm0.3$} & \textit{2.2} & 39.9 & \textit{30.8}\tiny{$\pm1.0$}\\
\hline
\end{tabular}
\vspace*{1mm}
\caption{We compare the proposed \textit{acc-tail} metric with other benchmarks. Results computed on the testdev split of \emph{GQA-OOD} and GQA~\cite{hudson2019gqa}, the test split of VQA-CP2~\cite{vqa-cp} and the VQA2~\cite{goyal2017making} validation split. Values in italic: trained and tested by ourselves.}
}
\end{table}
\setlength{\tabcolsep}{1.4pt}

\subsection{Comparison with other benchmarks}
\label{sub:compare}
\noindent
We compare the proposed \emph{GQA-OOD} benchmark with the following three standard VQA datasets:

\textbf{GQA~\cite{hudson2019gqa}} (balanced version) is a dataset with $\sim$1.7M question-answer pairs automatically generated from real images. Among all VQA datasets, GQA has the richest annotations allowing to evaluate models with complementary metrics: consistency, validity, plausibility, \cite{hudson2019gqa} \etc. Here, we only discuss overall accuracy and the distribution score on the GQA testdev split as the other metrics are unrelated to the topic of our paper. 
The distribution score is obtained by measuring the match between the true GT answer distribution and the  predicted distribution. 
      \textbf{VQA2~\cite{goyal2017making}} (265K images, ${\ge}3$ questions each) is composed of questions created by humans,
each question is annotated with 10 GT answers. We compare with overall accuracy on the VQA2 validation split.
     \textbf{VQA-CP2~\cite{vqa-cp}} has been constructed by reorganising the training and validation splits of 
VQA2~\cite{goyal2017making} aiming at differences in answer distributions between training and test splits.
It has been designed to measure sensitivity to language bias.

\myparagraph{Comparison with GQA and VQA2}
In Table~\ref{table:bench}, we compare our \textit{acc-tail} score with the other benchmarks. We can see that overall accuracy on GQA and VQA2 is not sufficient to fully characterize the VQA performances.
Our evaluation in OOD settings is the only one to reveal that even SOTA models struggle on infrequent question-answer pairs.
The best-performing model LXMERT looses about 10 points in the OOD setting. Our metric also unveils that, despite performing on-par with LXMERT on GQA overall, MMN struggles more on infrequent question-answer pairs.
Finally, we argue that \textit{acc-tail} is easier to interpret than the error distribution measure defined in GQA.

\myparagraph{Comparison with VQA-CP2~\cite{vqa-cp}}
Comparing \textit{acc-tail} to VQA-CP2 overall accuracy, we observe similar scores on standard VQA models,  
but a completely different behavior for bias-reduction methods. While they do not improve the scores in the OOD setting (\textit{cf.} Section~\ref{sub:err_dist}), they achieve strong performances on VQA-CP2.
The score of LM stands out, achieving the highest overall accuracy on VQA-CP2 ($52.0\%$) but one of the lowest \textit{acc-tail} on \emph{GQA-OOD} ($33\%$), with similar behavior for RUBi and BP. In short, while VQA-CP2 measures to what extent a VQA model struggles to generalize to a specific unseen distributions, the VQA-CP2 evaluation does not reflect the model behaviour on rare question-answer pairs. 

\section*{Discussion and conclusions}
\noindent
Going beyond previous attempts to reduce the influence of dataset biases in VQA evaluation, our proposed \textit{GQA-OOD} benchmark allows to \textit{both} evaluate (1) whether models have absorbed tendencies in the training data, and (2) how well they generalize to rare/unseen question-answer pairs.
This was made possible by (i) a thorough choice of imbalanced question groups, (ii) a new set of metrics and finally, (iii) by allowing to control the amount of distribution shift via the hyper-parameter $\alpha$. 
We have provided evidence that the benchmark and metric measure performance and dependency on dataset bias. 
Our experiments have also shown that neither conventional SOTA VQA models nor  dedicated bias reduction methods succeed in all aspects of the proposed evaluation benchmark. We hope that this sheds light on the current shortcomings in vision and language reasoning, and we hope that \textit{GQA-OOD} will contribute to the emergence of new models, less prone to learning spurious biases and more reliable in real-world scenarios.

{\small
\textbf{Acknowledgements ---} C. Wolf acknowledges support from ANR through grant ``\emph{Remember}'' (ANR-20-CHIA-0018).
}

{\small
\bibliographystyle{ieee_fullname}
\bibliography{main}
}

\newpage
\clearpage
\appendix

\section{Additional examples from the GQA-OOD validation split}

\begin{figure*}[] \centering
\begin{tabular}{cc}
    \begin{minipage}{4cm}
        \includegraphics[width=\linewidth]{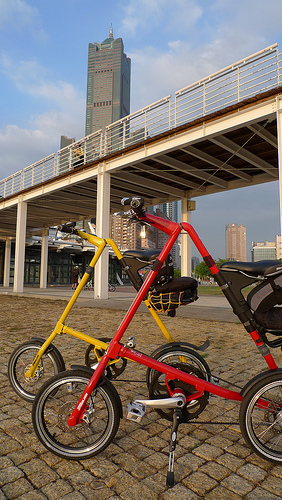}
    \end{minipage}
    &
    \begin{minipage}{12cm}
        \includegraphics[width=\linewidth]{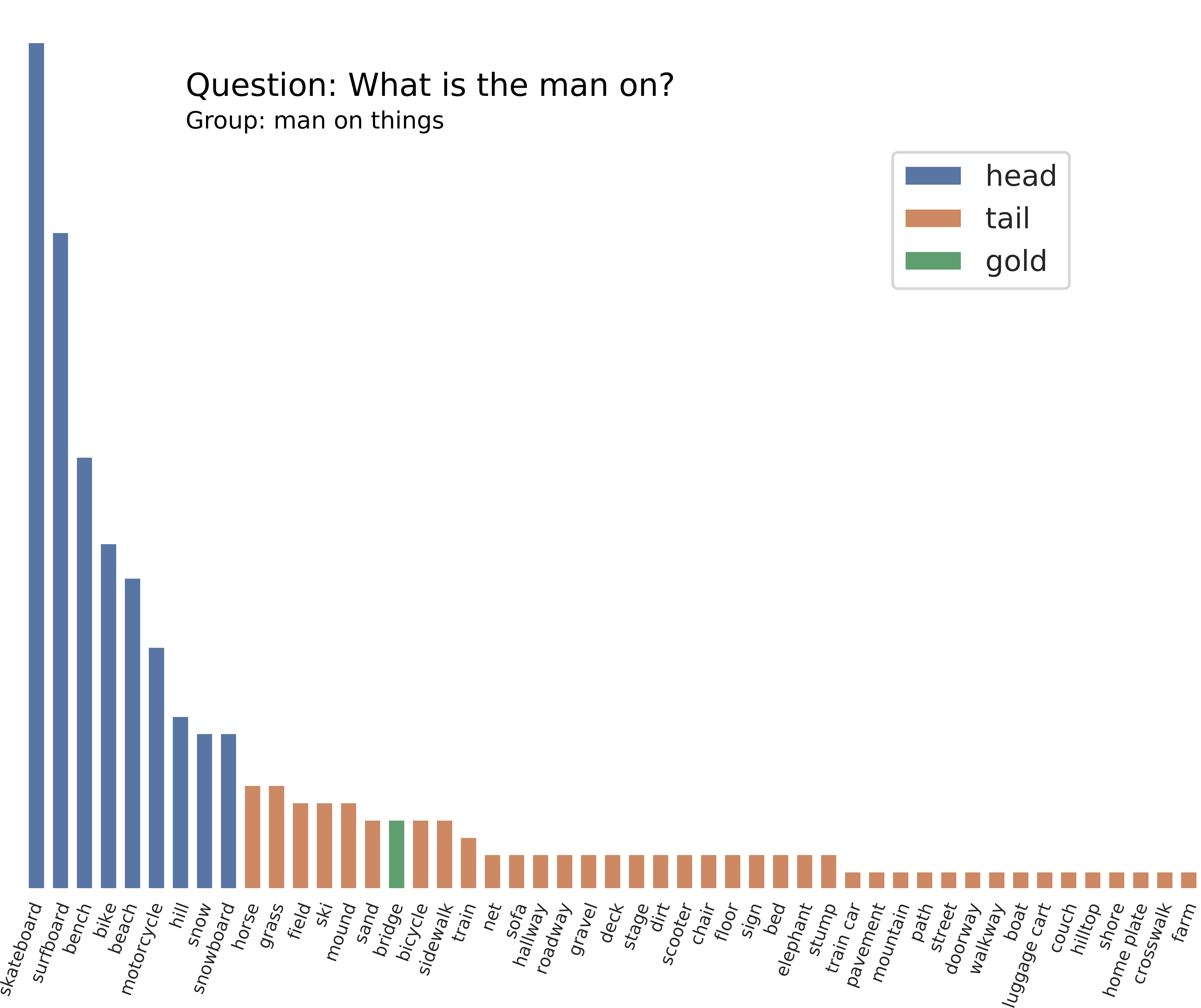}
    \end{minipage}
    \\
\end{tabular}
\vspace*{1mm}
\caption{\label{fig:bridge}Tail sample from the GQA-OOD validation split. Question:\textit{What is the man on?}. Answer:\textit{bridge}. The evaluated models have predicted:
LSTM=\textit{skateboard};
BUTD~\cite{anderson2018bottom}, MCAN~\cite{yu2019deep} = \textit{bike};
BAN~\cite{kim2018bilinear}, BUTD+LM~\cite{clark2019don}, MMN~\cite{chen2019meta}, BUTD+RUBI~\cite{cadene2019rubi}, BUTD+BP~\cite{clark2019don} = \textit{bicycle};
LXMERT~\cite{tan2019lxmert}, ORACLE-VIS = \textit{bridge}. The histogram represents the answer frequency measured over the set of all questions belonging to the question group.
}
\end{figure*}

\begin{figure*}[] \centering
\begin{tabular}{cc}
    \begin{minipage}{4cm}
        \includegraphics[width=\linewidth]{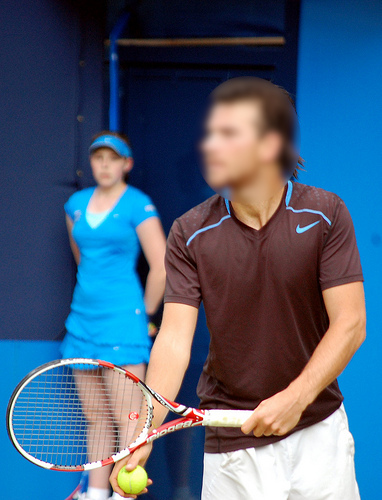}
    \end{minipage}
    &
    \begin{minipage}{9cm}
        \includegraphics[width=\linewidth]{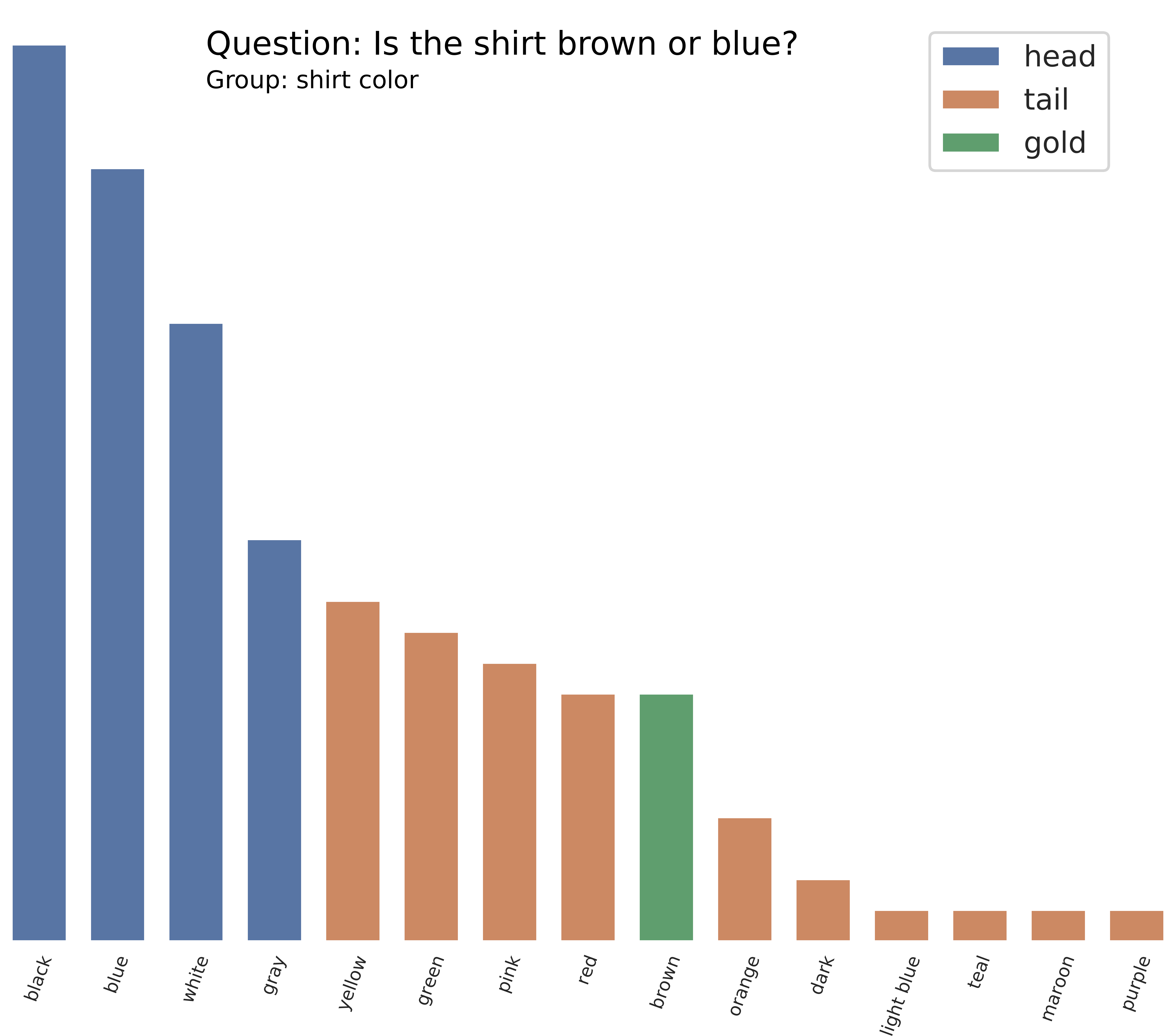}
    \end{minipage}
    \\
\end{tabular}
\vspace*{1mm}
\caption{\label{fig:shirt}Tail sample from the GQA-OOD validation split. Question:\textit{Is the shirt brown or blue?}. Answer:\textit{brown}. The evaluated models have predicted:
LSTM, BAN~\cite{kim2018bilinear}, BUTD~\cite{anderson2018bottom}, BUTD+LM~\cite{clark2019don} = \textit{blue};
BUTD+RUBI~\cite{cadene2019rubi}, = \textit{light blue};
MCAN~\cite{yu2019deep}, LXMERT~\cite{tan2019lxmert}, ORACLE-VIS, MMN~\cite{chen2019meta}, BUTD+BP~\cite{clark2019don} = \textit{brown}. The histogram represents the answer frequency measured over the set of all questions belonging to the question group.
}
\end{figure*}

\begin{figure*}[] \centering
\begin{tabular}{cc}
    \begin{minipage}{6cm}
        \includegraphics[width=\linewidth]{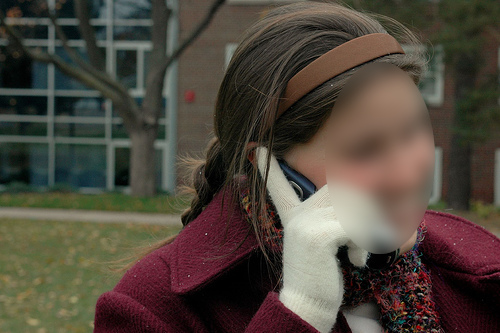}
    \end{minipage}
    &
    \begin{minipage}{9cm}
        \includegraphics[width=\linewidth]{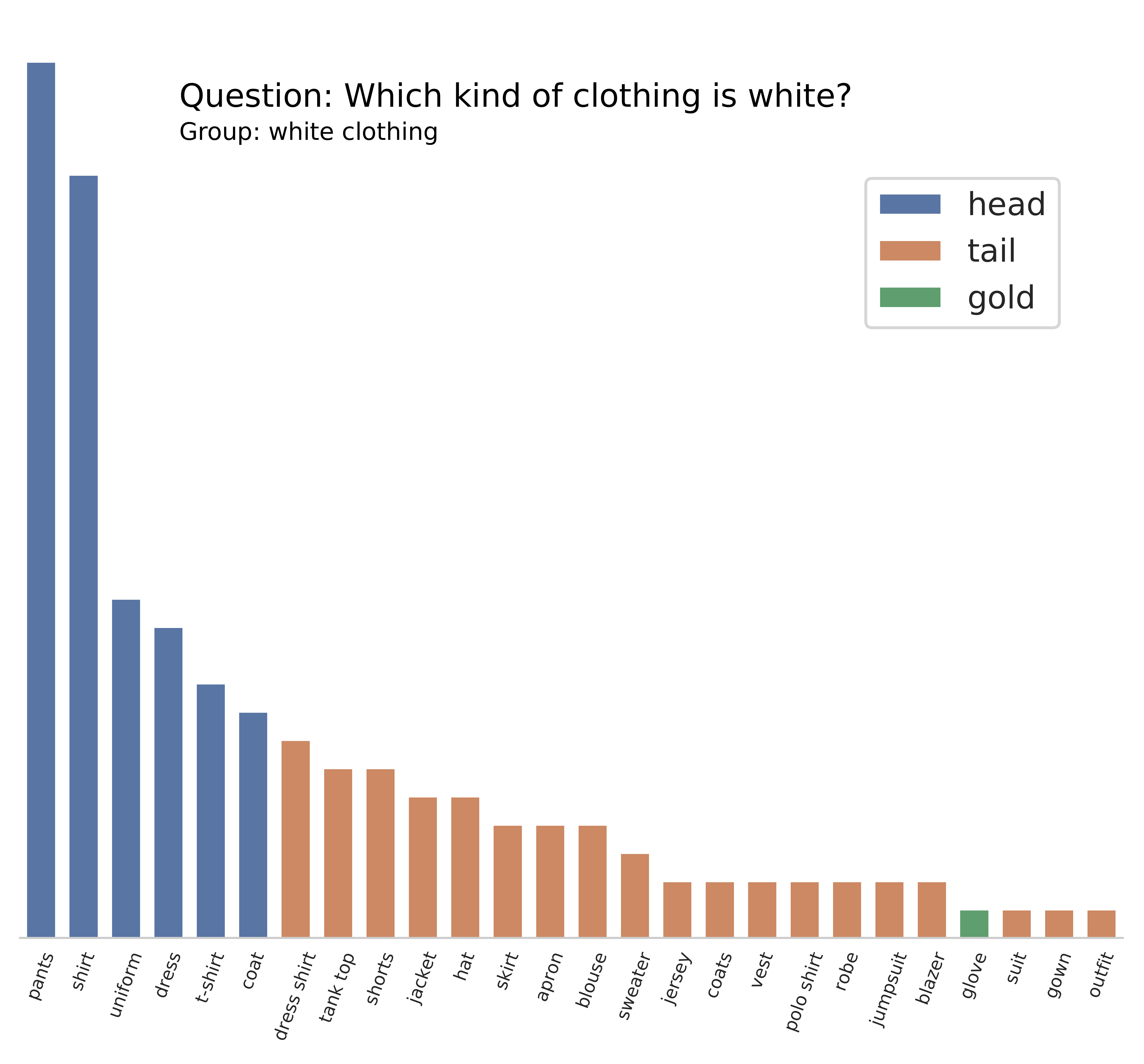}
    \end{minipage}
    \\
\end{tabular}
\vspace*{1mm}
\caption{\label{fig:glove}Tail sample from the GQA-OOD validation split. Question:\textit{Which kind of clothing is white?}. Answer:\textit{glove}. The evaluated models have predicted:
LSTM = \textit{shirt};
LXMERT~\cite{tan2019lxmert}, BUTD~\cite{anderson2018bottom}, BAN~\cite{kim2018bilinear}, MMN~\cite{chen2019meta}, BUTD+RUBI~\cite{cadene2019rubi} = \textit{coat};
MCAN~\cite{yu2019deep},= \textit{jacket};
BUTD+LM~\cite{clark2019don}, BUTD+BP~\cite{clark2019don}= \textit{long sleeved};
ORACLE-VIS = \textit{glove}. The histogram represents the answer frequency measured over the set of all questions belonging to the question group.
}
\end{figure*}

\begin{figure*}[] \centering
\begin{tabular}{cc}
    \begin{minipage}{5cm}
        \includegraphics[width=\linewidth]{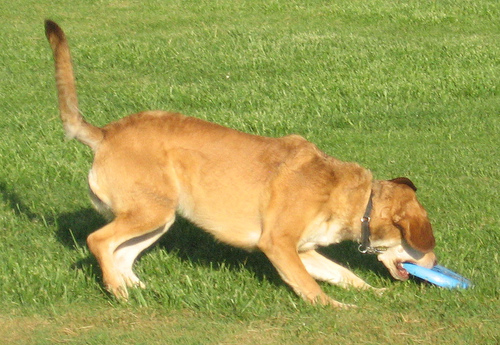}
    \end{minipage}
    &
    \begin{minipage}{9cm}
        \includegraphics[width=\linewidth]{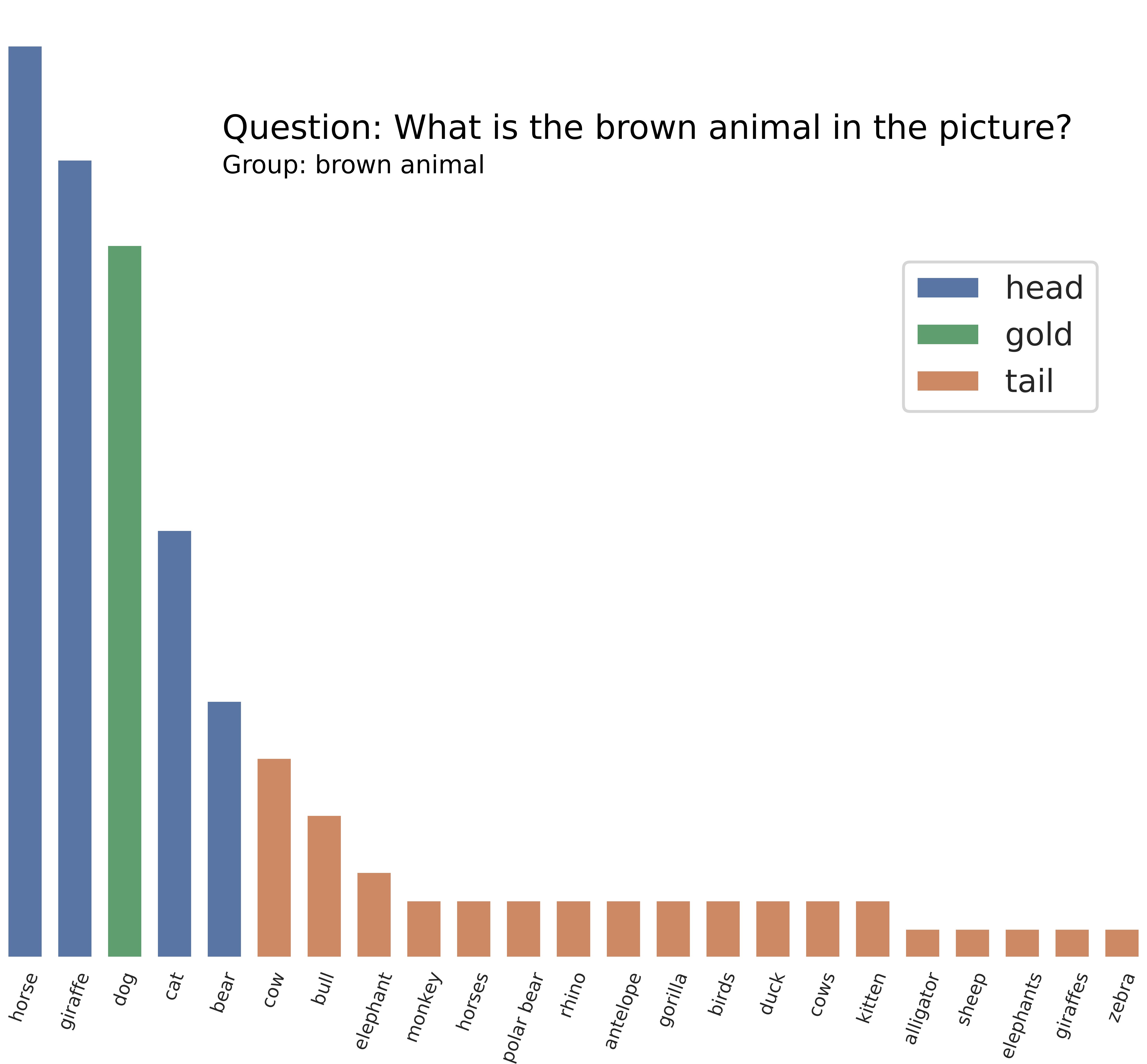}
    \end{minipage}
    \\
\end{tabular}
\vspace*{1mm}
\caption{\label{fig:dog}Head sample from the GQA-OOD validation split. Question:\textit{What is the brown animal in the picture?}. Answer:\textit{dog}. The evaluated models have predicted:
LSTM = \textit{horse};
BAN~\cite{kim2018bilinear}, BUTD~\cite{anderson2018bottom}, BUTD+LM~\cite{clark2019don}, BUTD+RUBI~\cite{cadene2019rubi}, MCAN~\cite{yu2019deep}, LXMERT~\cite{tan2019lxmert}, ORACLE-VIS, MMN~\cite{chen2019meta}, BUTD+BP~\cite{clark2019don} = \textit{dog}. The histogram represents the answer frequency measured over the set of all questions belonging to the question group.
}
\end{figure*}

In order to give a better insight about the benchmark's goals and possibilities, we provide additional samples extracted from the GQA-OOD validation split.
In Figure~\ref{fig:bridge} and \ref{fig:shirt}, we show two question-answer pairs belonging to the tail. The histogram represents the answer frequency measured over the set of all questions belonging to the group of the given question. We colored the answers according to their label, head or tail. First, we can observe that the histogram is very imbalanced, which motivates the GQA-OOD approach. Second, in the caption we provide the predicted answer for each one of the evaluated model. One can notice that the predictions are diverse, showing various degree of bias dependency.
However, all models are mostly relying on context biases, as shown in Figure~\ref{fig:glove}.
Finally, in Figure~\ref{fig:dog}, we show a question-answer pair labelled as head, where all models (excepted the blind LSTM) are correct.

\section{Dataset statistics}
\label{sec:appendix_statistics}

We provide some analysis and statistics to assess the reliability of the proposed benchmark. In particular, we analyse the nature and the distribution of the questions involved in \emph{GQA-OOD} and demonstrate that it preserves the original question diversity of GQA~\cite{hudson2019gqa}. 

\begin{figure}[]
    \centering
    \begin{minipage}{0.4\textwidth}
        \centering
        \includegraphics[width=0.9\linewidth]{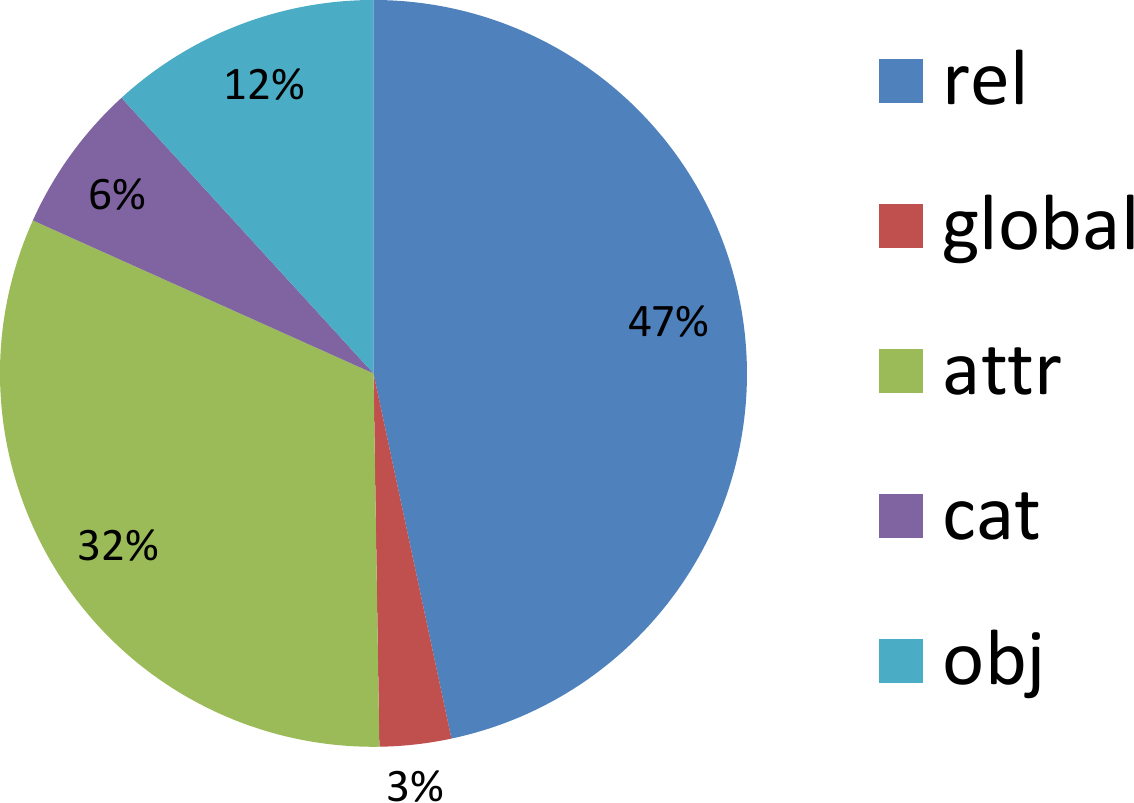}
        \caption{Distribution of the semantic types in GQA.}
        \label{fig:sem_val}
    \end{minipage}\hfill
    \begin{minipage}{0.4\textwidth}
        \centering
        \includegraphics[width=0.9\linewidth]{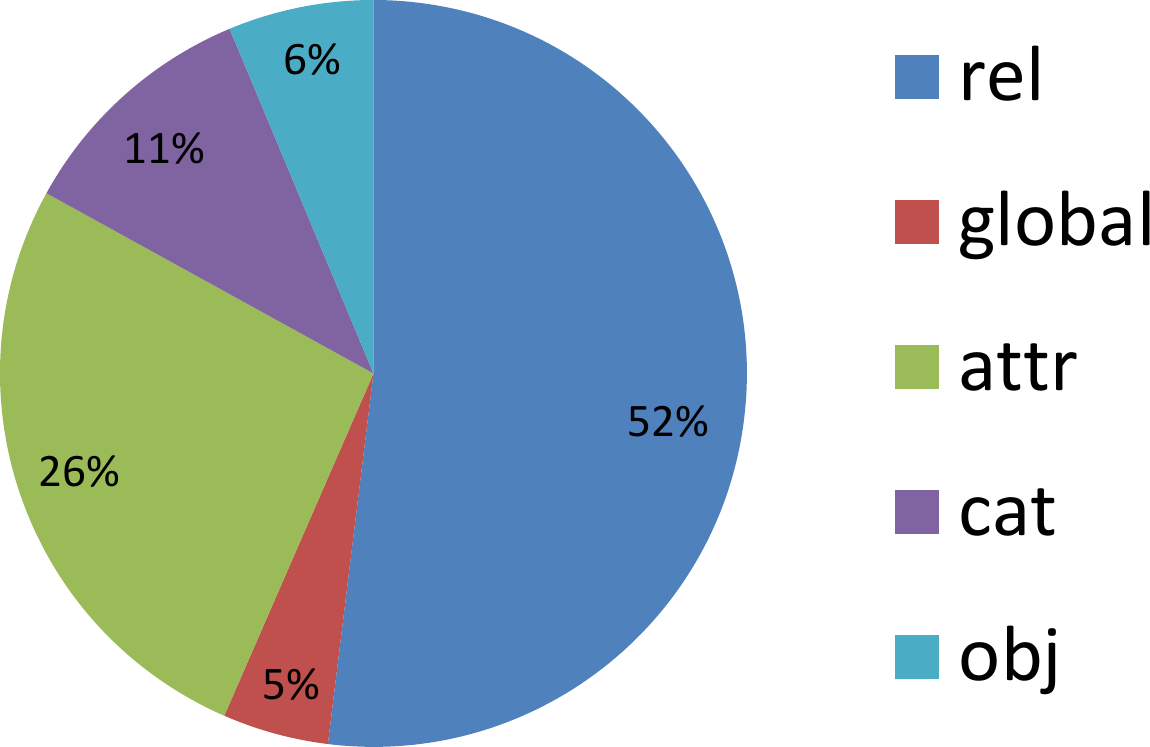}
        \caption{Distribution of the semantic types in \emph{GQA-OOD} (tail).}
        \label{fig:sem_ood}
    \end{minipage}
\end{figure}

\begin{figure}[]
    \centering
    \begin{minipage}{0.4\textwidth}
        \centering
        \includegraphics[width=0.9\linewidth]{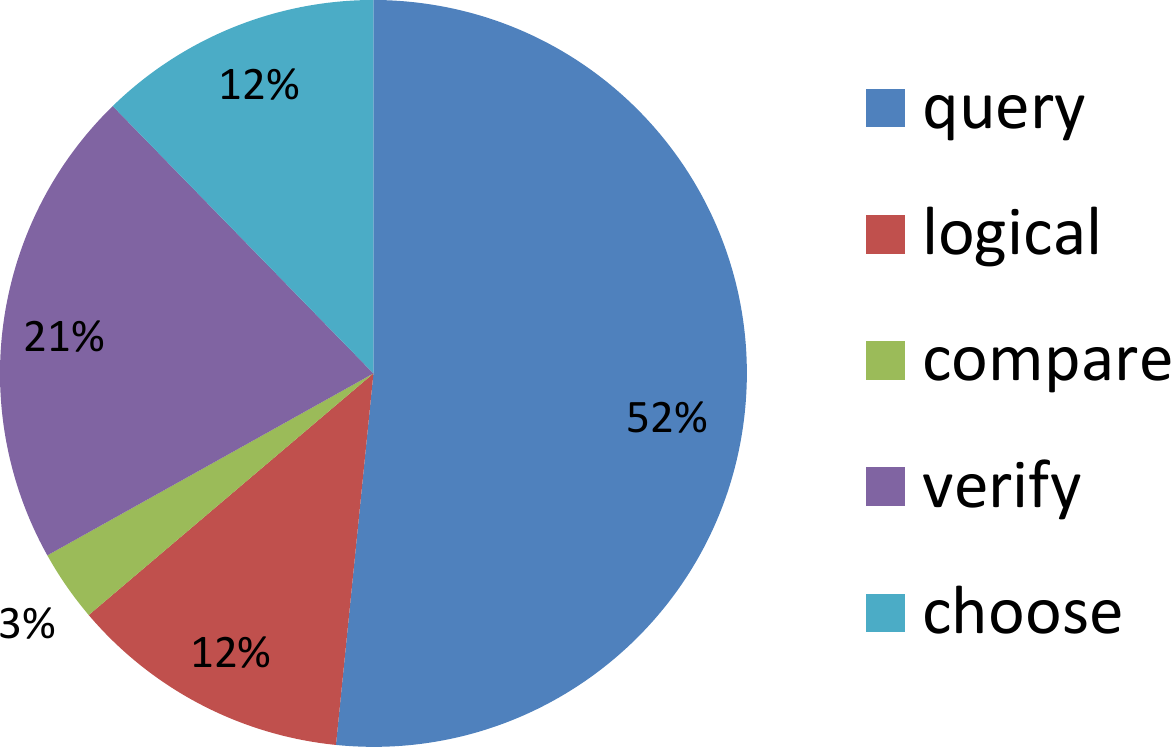}
        \caption{Distribution of the structural types in GQA.}
        \label{fig:struc_val}
    \end{minipage}\hfill
    \begin{minipage}{0.4\textwidth}
        \centering
        \includegraphics[width=0.9\linewidth]{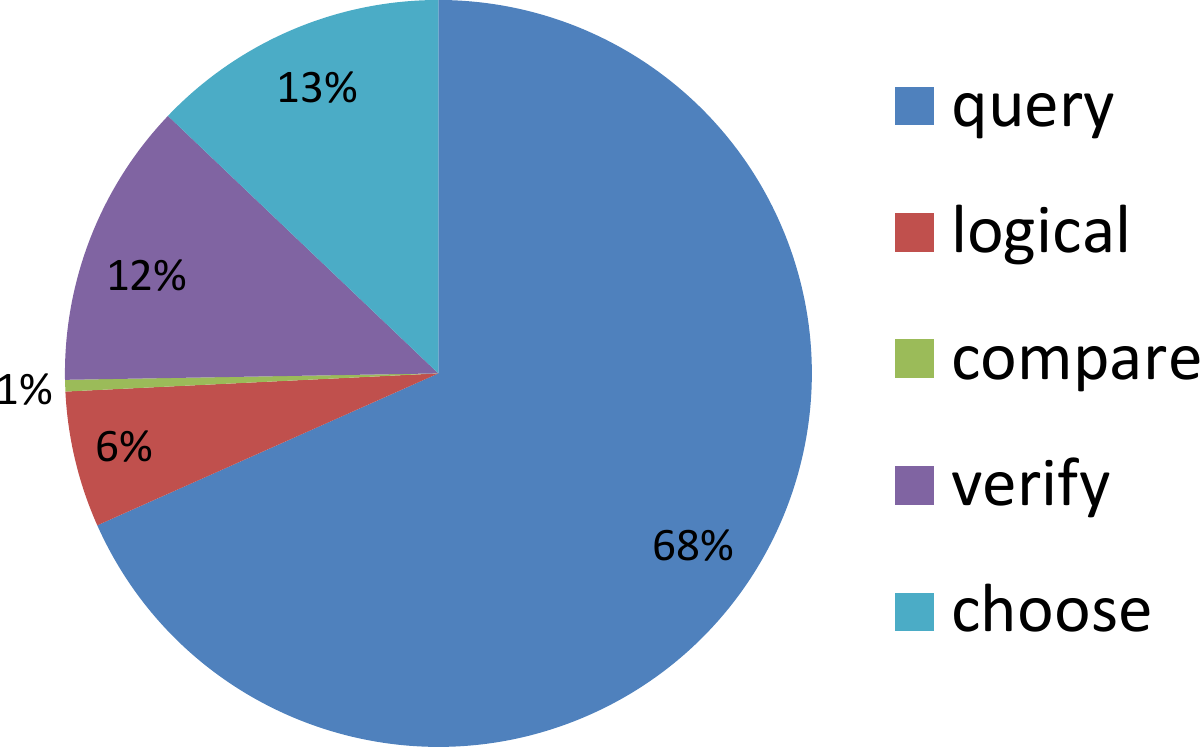}
        \caption{Distribution of the semantic types in \emph{GQA-OOD} (tail).}
        \label{fig:struc_ood}
    \end{minipage}
\end{figure}

\myparagraph{Question diversity}
\label{A:diversity}
Figure~\ref{fig:struc_val} and Figure~\ref{fig:struc_ood} show the distribution of question structure type as defined in GQA~\cite{hudson2019gqa} on the validation split. As one can observe, the process implemented to construct \emph{GQA-OOD} does not alter the question diversity of the original split. However, the proportion of open questions -- 'query' in Figure~\ref{fig:sem_val} and Figure~\ref{fig:sem_ood} -- has increased in \emph{GQA-OOD}. Indeed, open questions -- such as color questions -- generally accept a wider diversity of answer, therefore it is prone to be more imbalanced. At contrary, other types such as `choose', `verify' or `compare' usually accept only two possible answers and are easier to balance. Figure~\ref{fig:sem_val} and Figure~\ref{fig:sem_ood} details the distribution of the structure types in the validation in \emph{GQA-OOD} compared to GQA. 

\section{Training details}
\label{sec:training}

\myparagraph{Training hyper-parameters}
All models evaluated on GQA and \emph{GQA-OOD} have been trained on the balanced training set of GQA, and validated on the validation split. For MCAN and BUTD we use publicly available implementations at \url{https://github.com/MILVLG/openvqa}.
LSTM, BUTD~\cite{anderson2018bottom}, RUBi~\cite{cadene2019rubi}, BP~\cite{clark2019don} and LM~\cite{clark2019don} are trained during 20 epochs with a batch size equals to $512$ and Adam~\cite{kingma2014adam} optimizer. At the beginning of the training we linearly increase the learning rate from $2e^{-3}$ to $2e^{-1}$ during 3 epochs, followed by a decay by a factor of $0.2$ at epochs 10 and 12.
MCAN~\cite{yu2019deep} is trained during 11 epoch with a batch size equals to $64$ and Adamax~\cite{kingma2014adam} optimizer. At the beginning of the training we linearly increase the learning rate from $1e^{-4}$ to $2e^{-1}$ during 3 epochs, followed by a decay by a factor of $0.2$ at epochs 10 and 12.
For MMN~\cite{chen2019meta}, we use the author's implementation and trained model available at \url{https://github.com/wenhuchen/Meta-Module-Network}.

\myparagraph{LXMERT pre-training}
LXMERT~\cite{tan2019lxmert} is pre-trained on a corpus gathering images and sentences from MSCOCO~\cite{lin2014microsoft} and VisualGenome~\cite{krishna2017visual}. As the GQA dataset is built upon VisualGenome, the original LXMERT pre-training dataset contains samples from the GQA validation split. Hence, we remove those samples before pre-training in order to correctly evaluate on the GQA and GQA-OOD validation split.

\myparagraph{Visual Oracle}
The VIS-ORACLE model is based on a tiny version of the LXMERT architecture~\cite{tan2019lxmert}, where we set the hidden size to $128$ and the number of per-layer heads to $4$.
This perfect-sighted model is taken as input objects extracted from the ground-truth GQA annotation~\cite{hudson2019gqa}. Each object is constructed using one hot vectors encoding its class, its attributes and its in and out scenegraph relationships.

\begin{figure}[h]
    \centering
    \includegraphics[width=1\linewidth]{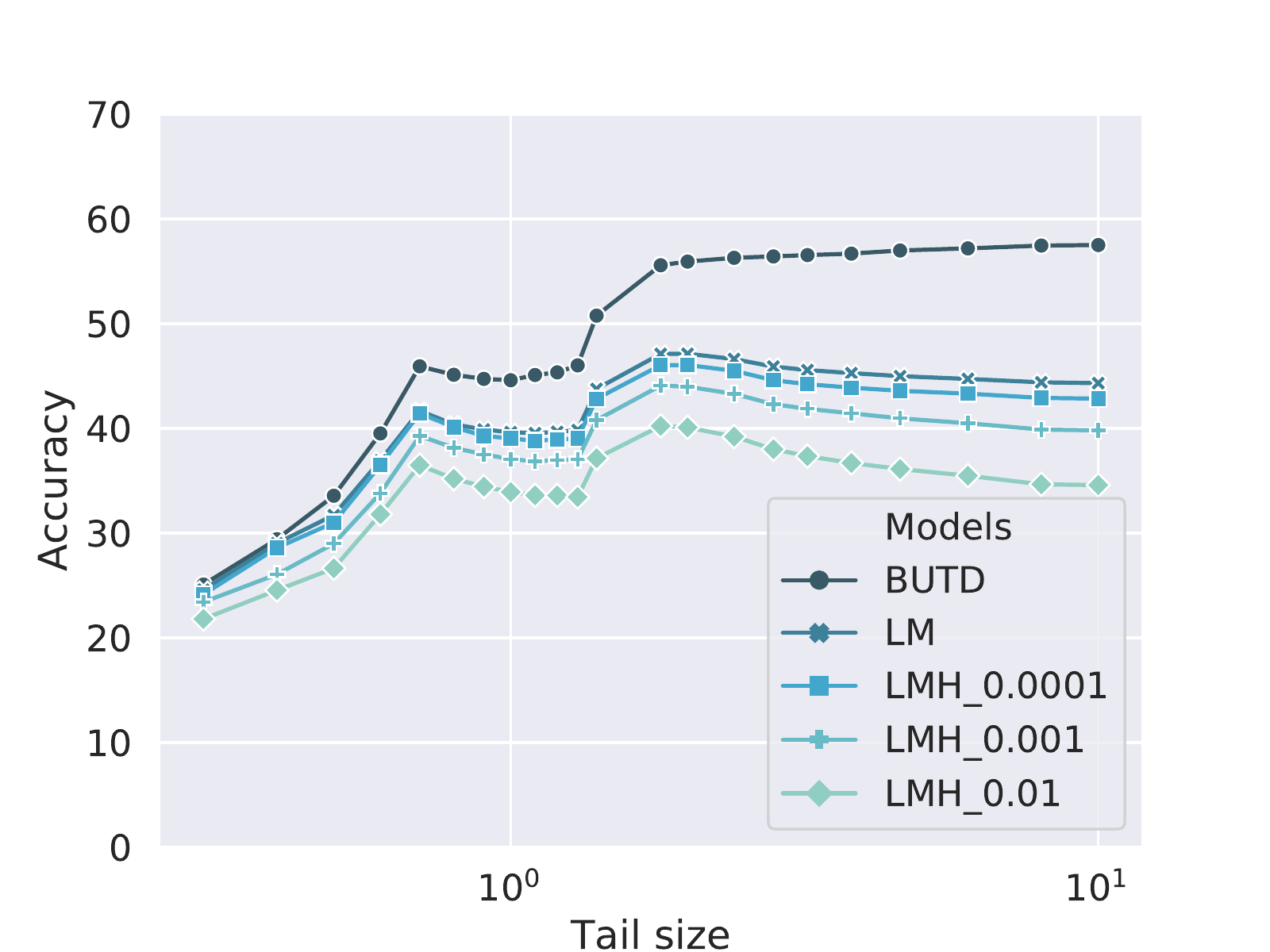}
    \caption{Influence of the LM entropy penalty weight on the prediction error distribution.}
    \label{fig:tail_lmh}
\end{figure}

\myparagraph{LM hyper-parameters}
Figure~\ref{fig:tail_lmh} details the hyper-parameter search for the entropy penalty weight in LM~\cite{clark2019don}. We found that the entropy penalty was degrading the \emph{GQA-OOD} accuracy when training on GQA. In particular, the flattening of the right side of the curve (most frequent samples) is even more present for higher penalty weight.

\section{Measuring head/tail confusion}
In the paper, we measure the head/tail confusion to get an insight on what extent the prediction errors are related to a context bias dependency. For the sake of clarity, we omit the detailed description of this procedure in the main paper. Nevertheless, the reader can find the exact methodology in the following paragraph.

The confusion corresponds to the proportion of questions where the model predicts a \textit{head} answer with a \textit{tail} GT answer. When plotting the confusion versus $\alpha$, we decrease the size of the tail set (\ie{} we keep only the rarest question-answer pairs) while keeping the head set unchanged. Then we observe that for the majority of models, the rarest the GT answer is, the more probable the prediction belongs to the head.

\end{document}